\pdfoutput=1
\documentclass{article}

\PassOptionsToPackage{dvipsnames,table}{xcolor}

\usepackage{microtype}
\usepackage{graphicx}
\usepackage{booktabs} %

\usepackage{hyperref}
\usepackage{url}

\usepackage[final,main]{neurips_2026}
\makeatletter
\renewcommand{\paragraph}{\@startsection{paragraph}{4}{\z@}{0.35ex \@plus 0.2ex \@minus 0.1ex}{-1em}{\normalsize\bf}}
\makeatother
\usepackage{amsmath}
\usepackage{amssymb}
\usepackage{amsfonts}

\let\oldsubsection\subsection
\renewcommand{\subsection}{\vspace{0.15em}\oldsubsection}

\usepackage{xcolor}
\usepackage{multirow}
\usepackage{qtree}
\usepackage{multicol}
\usepackage{threeparttable}
\usepackage{tabularx}
\usepackage{longtable}
\usepackage{dcolumn}
\newcolumntype{d}[1]{D..{#1}} %
\usepackage{nicefrac}

\usepackage[capitalize,noabbrev]{cleveref}

\makeatletter
\newenvironment{apptable}{\par\vspace{\intextsep}\noindent\begin{minipage}{\linewidth}\def\@captype{table}\setlength{\abovecaptionskip}{0pt}\setlength{\belowcaptionskip}{7pt}}{\end{minipage}\par\vspace{\intextsep}}
\makeatother

\definecolor{darkpowderblue}{rgb}{0.0, 0.05, 0.5}
\definecolor{ForestGreen}{rgb}{0.13, 0.55, 0.13}
\hypersetup{
 colorlinks,
 pdftitle={Ordinary Least Squares as an Attention Mechanism},
 pdfauthor={Philippe Goulet Coulombe},
 citecolor=darkpowderblue,
 linkcolor=red,
 urlcolor=blue}

\title{Ordinary Least Squares as an Attention Mechanism}

\author{%
 Philippe Goulet Coulombe \\
 Universit\'e du Qu\'ebec \`a Montr\'eal \\
 \texttt{p.gouletcoulombe@gmail.com}
}

\begin{document}

\maketitle

\begin{abstract}
I show that ordinary least squares (OLS) predictions can be rewritten as the output of a restricted attention module, akin to those forming the backbone of large language models. The connection comes from viewing OLS as a similarity-based prediction rule in a learned embedding space. In this representation, least squares does not estimate coefficients per se. Instead, it selects an embedding that minimizes squared prediction error by matching training and test vectors through inner products. This maps directly onto the query-key-value structure of attention mechanisms. I then discuss extensions to dimensionality reduction, nonlinearity, and time series econometrics. Monte Carlo simulations and real-data experiments on UCI/OpenML benchmarks show that nonlinear Attention Regression performs competitively against standard machine learning baselines. In the reverse direction, I replace the attention sublayer of a transformer for tabular data with an explicit regression on polynomial features. The resulting model performs comparably to the standard transformer at a fraction of its parameter count.
\end{abstract}

\section{Introduction}

It seems fair to say that the self-attention mechanism of \cite{vaswani2017attention} sparked a technical revolution. Its introduction opened the era of large language models (LLMs), moving natural language processing from convolutional and recurrent architectures to Transformers.

Yet, for readers trained in applied statistics, the query--key--value formulation of attention, borrowed from information retrieval, may be elusive. In this paper, I show that test-set predictions from a linear regression estimated via ordinary least squares (OLS) can be rewritten as the output of a simplified attention module, where the weight matrices admit a closed-form solution. As it turns out, the connection is useful for understanding both, and it takes the mystery out of attention for anyone familiar with ordinary least squares.

\begin{figure}[t]
  \centering
  \includegraphics[width=\textwidth]{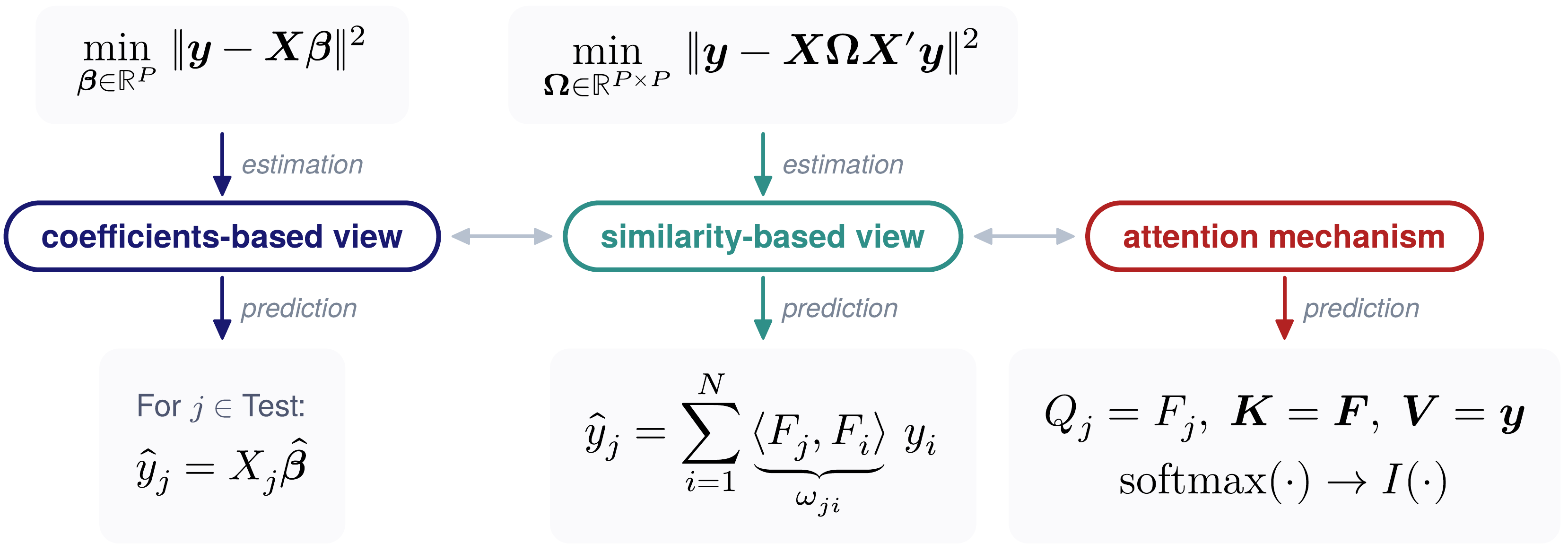}
  \caption{Roadmap: Three equivalent views of OLS predictions. \textbf{Left:} Coefficient-based formulation. \textbf{Middle:} Similarity-based interpretation---predictions as weighted averages of training outcomes in a transformed space. \textbf{Right:} Restricted attention module with identity activation.}
  \label{fig:three_views}
\end{figure}

\noindent \textbf{The Double Life of Least Squares.} Beneath its standard interpretation, OLS leads a double life. Each prediction is a similarity-weighted average of training outcomes, where the weights are inner products in an orthonormalized predictor space; observations closer to the target receive more \textit{attention}. Hence, OLS is as much proximity-based as it is correlation-based. This reading turns the eigendecomposition of $(\boldsymbol{X}'\boldsymbol{X})^{-1}$ into the optimal embedding, the unique inner-product similarity metric that minimizes squared prediction error for every outcome vector. Were one to apply gradient descent to a linear attention module trained to predict $y$, its predictions would converge to exactly those of OLS. The headline result of Section~\ref{sec2} is then that OLS predictions are exactly a restricted attention module (Figure~\ref{fig:three_views}). The paper does not claim that LLMs perform OLS; it claims that the principles of attention are closer to least squares than is usually assumed.

\noindent The equivalence is also useful in both directions. If attention is regression, nonparametric regression knowledge (initialization, regularization, covariance estimation) transfers directly to attention modules. Taking the formula at face value, the resulting ``Attention Regression'' estimator lands in the neighborhood of random forests and MLPs on UCI/OpenML benchmarks (Section~\ref{sec:exp}). The reverse direction asks whether the attention block inside a transformer can itself be replaced by an explicit regression on expanded features. Section~\ref{sec:exp} reports a ``Regression Block'' that replaces the attention sublayer with an explicit polynomial regression and performs comparably to the FT-Transformer, ahead on all eight benchmarks by small margins, at about thirty percent of the parameters. The ablations of Appendix~\ref{app:ablations} identify the explicit regression on polynomial features as the component responsible.

\noindent \textbf{Roadmap.} Section~\ref{sec2} establishes the identity. Section~\ref{sec25} revisits dimensionality reduction, shrinkage, nonlinearity and depth in the same terms. Section~\ref{sec:exp} puts the equivalence to work empirically in both directions, and the appendices carry the time-series connections and the ablations.

\noindent \textbf{Related Work.} \textit{Kernels and smoothers.} The connection between attention mechanisms---based on a nonlinearly transformed dot product in a latent space---and kernel methods has been widely discussed in the literature \citep{tsai2019transformer, choromanski2020rethinking, peng2021random}. However, the parallel runs from attention toward kernel smoothers, which are nonparametric regressions in their own right, and stops short of a correlation-based estimator. This paper closes the loop from the other side: even unmodified OLS---no kernel, no softmax---is itself an attention module in disguise.

From the OLS predictions perspective, perhaps the closest formulation to the one presented in this paper is that of the representer theorem, which states that solutions to nonparametric ridge regression can be expressed as a finite combination of kernelized inner products between the new data vector and training samples \citep{scholkopf2001generalized,wahba2019representer}. However, these formulations always feature a vector of observation-specific weights which serves as a set of parameters defining the dual solution to regularized least squares problems. The precise interpretation of the parameters' role, beyond serving as linear weights for kernelized inner products in the dual representation of predictions, remains unclear.

In this paper's formulation, where OLS is reframed as an attention module, weights on training observations emerge directly from comparing training and test data in a transformed, orthonormal space. This transformation enables the direct assignment of weights to training outcomes, analogous to “values” in the information retrieval framework.

\noindent \textit{Attention on regression tasks.} Two recent studies have drawn connections between attention mechanisms and regression tasks. \cite{marion2024attention} introduced the single-location regression task, a simplified setting where only one token in a sequence determines the output, with its position treated as a latent variable. They demonstrated that a non-linear self-attention layer can effectively identify the relevant token and perform the particular regression. \cite{garnelo2023exploring} expanded on this by examining key-value-query models, which share the input structure of attention mechanisms but differ in computation. They introduced a new module that generalizes linear attention and computes regularized least squares solutions within the key-value-query framework. The present paper focuses on the standard attention mechanism and explains why similarities between regression solutions and attention modules arise: least squares-based predictions are also interpretable as inner-product-based similarity computations in an optimized orthonormal space--an interpretation closely aligned with how keys, queries, and values interact in LLMs.   

\noindent \textit{What trained attention learns.} The closest work is the in-context-learning literature, which asks what a Transformer trained on regression prompts learns. \citet{garg2022what} show that it learns to behave like least squares. \citet{akyurek2023what} and \citet{vonoswald2023transformers} show by construction that attention layers can implement gradient-descent updates and, with enough depth and width, recover ridge or least squares. \citet{mahankali2024one} and \citet{ahn2023transformers} characterize the loss-minimizing parameters of one layer of linear attention, pre-trained over random regression tasks, as one preconditioned gradient step, the preconditioner tied to the inverse population covariance, and \citet{zhang2024trained} show that gradient flow reaches that optimum. \citet{tarzanagh2023transformers} supply the closest reading from the training side: the optimization geometry of self-attention is that of a hard-margin SVM separating optimal from non-optimal tokens, so their result concerns token selection and the regression result here concerns how the selected values are combined.

\noindent This paper runs in the \textit{reverse direction}. Rather than training a Transformer and asking which regression algorithm it learns, it starts from the least-squares prediction rule and asks what attention operation it already contains: conditional on the observed design matrix, least-squares prediction \emph{is} a restricted attention operation with embedding $(\boldsymbol{X}'\boldsymbol{X})^{-1}$. In the work above the embedding is learned across many tasks and is a population-level quantity; here it is the sample-specific inverse Gram matrix, supplied in closed form by the estimator. After all, this is not surprising: a rule that already has an attention form is a natural thing for attention to learn.

\section{An Equivalence}\label{sec2}

Figure~\ref{fig:three_views} provides a roadmap of the equivalence established in this section. I start with the standard OLS prediction (left panel), then introduce a less conventional but illuminating interpretation---OLS as a similarity-based retrieval process in a transformed space (middle panel). I then show this maps directly onto the query-key-value structure of attention mechanisms (right panel).

\subsection{A Proximity Interpretation of Least Squares}\label{prox_ols}

Let \( \boldsymbol{X}_{\text{train}} \in \mathbb{R}^{N \times P} \) denote the training design matrix and \( \boldsymbol{y} \in \mathbb{R}^{N} \) the corresponding response vector, where in-sample observations run from \( i = 1 \) to \( N \). The OLS estimator is given by
\begin{align}
  \hat{\boldsymbol{\beta}} &= (\boldsymbol{X}_{\text{train}}' \boldsymbol{X}_{\text{train}})^{-1} \boldsymbol{X}_{\text{train}}' \boldsymbol{y}.
\end{align}
For out-of-sample observations indexed by \( j = N+1, \dots, N+J \), the corresponding feature matrix is \( \boldsymbol{X}_{\text{test}} \in \mathbb{R}^{J \times P} \), and the predicted values are
\begin{align}\label{ols_standard}
  \hat{\boldsymbol{y}}_{\text{test}} = \boldsymbol{X}_{\text{test}} \hat{\boldsymbol{\beta}} = \boldsymbol{X}_{\text{test}} (\boldsymbol{X}_{\text{train}}' \boldsymbol{X}_{\text{train}})^{-1} \boldsymbol{X}_{\text{train}}' \boldsymbol{y}.
\end{align}
What is less widely recognized is that linear regression predictions can be interpreted as a proximity-based estimator. Using the eigendecomposition of \( (\boldsymbol{X}_{\text{train}}' \boldsymbol{X}_{\text{train}})^{-1} \), we rewrite the standard out-of-sample prediction formula \eqref{ols_standard} as
\begin{align}
\hat{\boldsymbol{y}}_{\text{test}} = \boldsymbol{X}_{\text{test}} \boldsymbol{U} \Lambda^{-1} \boldsymbol{U}' \boldsymbol{X}_{\text{train}}' \boldsymbol{y}_{\text{train}},
\end{align}
where \( \boldsymbol{U} \Lambda^{-1} \boldsymbol{U}' \) is the eigendecomposition of \( (\boldsymbol{X}_{\text{train}}' \boldsymbol{X}_{\text{train}})^{-1} \), with \( \Lambda \) being the diagonal matrix of eigenvalues of \( \boldsymbol{X}_{\text{train}}' \boldsymbol{X}_{\text{train}} \), and \( \boldsymbol{U} \) its eigenvectors. Decomposing \( \Lambda^{-1} \) into $\Lambda^{-\frac{1}{2}}\Lambda^{-\frac{1}{2}}$, we obtain:
\begin{align}
\hat{\boldsymbol{y}}_{\text{test}} = \underbrace{\boldsymbol{X}_{\text{test}} \boldsymbol{U} \Lambda^{-\frac{1}{2}}}_{\boldsymbol{F}_{\text{test}}} \underbrace{\Lambda^{-\frac{1}{2}} \boldsymbol{U}' \boldsymbol{X}_{\text{train}}'}_{\boldsymbol{F}_{\text{train}}'} \boldsymbol{y}_{\text{train}}.
\end{align}
This formulation highlights a natural similarity-based interpretation: both \( \boldsymbol{F}_{\text{test}} =\boldsymbol{X}_{\text{test}} \boldsymbol{U} \Lambda^{-\frac{1}{2}} \) and \( \boldsymbol{F}_{\text{train}} = \boldsymbol{X}_{\text{train}} \boldsymbol{U} \Lambda^{-\frac{1}{2}} \) are standardized factor scores: $\boldsymbol{F}_{\text{train}}$ is an orthonormal representation of $\boldsymbol{X}_{\text{train}}$, and $\boldsymbol{F}_{\text{test}}$ applies the same map to $\boldsymbol{X}_{\text{test}}$. The representation preserves the original input dimension, so $\boldsymbol{F}_{\text{train}} \in \mathbb{R}^{N \times P}$ and $\boldsymbol{F}_{\text{test}} \in \mathbb{R}^{J \times P}$. For $\boldsymbol{F}_{\text{train}}$, we have that
\begin{align}\label{ols_factor}
\boldsymbol{F}_{\text{train}}' \boldsymbol{F}_{\text{train}} = I_P.
\end{align}
However, since \(\Lambda\) and \(\boldsymbol{U}\) are estimated from the training data only, the identity $\boldsymbol{F}_{\text{test}}' \boldsymbol{F}_{\text{test}} = {I}_P $
does not hold exactly---just as \(\hat{\boldsymbol{\beta}}\) minimizes squared residuals by construction on the training data, but not necessarily on the test sample.

Thus, OLS out-of-sample predictions can be rewritten as:
\begin{align}\label{gram}
\underbrace{\hat{\boldsymbol{y}}_{\text{test}}}_{J \times 1} = \underbrace{\left[\boldsymbol{F}_{\text{test}} \boldsymbol{F}_{\text{train}}' \right]}_{J \times N} \, \, \underbrace{\boldsymbol{y}_{\text{train}}}_{N \times 1},
\end{align}
where \( \boldsymbol{F}_{\text{test}} \boldsymbol{F}_{\text{train}}' \) is a matrix of inner products between test observations (represented in \( \mathbb{R}^{P} \) by \( \boldsymbol{F}_{\text{test}} \)) and training observations (represented by \( \boldsymbol{F}_{\text{train}} \)). This reveals OLS as an inner-product-based similarity method, reinforcing its connection to traditional similarity-based estimation such as nearest neighbors. That is, the term \( \boldsymbol{F}_{\text{test}} \boldsymbol{F}_{\text{train}}' \) serves as a proximity matrix, measuring the similarity between pairs of training and test observations. Unlike the usual dual formula for ridge regression or likewise regularized problems, this view integrates the equivalent of dual coefficients $\hat{\boldsymbol{\alpha}} = ( \boldsymbol{X}_{\text{train}}\boldsymbol{X}_{\text{train}}'+\lambda I_N)^{-1}\boldsymbol{y}_{\text{train}} $ directly into the interpretation of test set predictions. %

\paragraph{Inner Product and Cosine Similarity Weighting.} The view of OLS as a similarity-based method becomes more evident when considering the equivalent formulation of equation \eqref{gram} for a single prediction of test sample observation \( j \):  
\begin{align}\label{innerprod}
\hat{y}_{j} &= \sum_{i=1}^{N} \underbrace{\langle F_{j}, F_i \rangle}_{\equiv \omega_{ji}} y_i.
\end{align}
Here, \(\langle \cdot , \cdot \rangle\) denotes the Euclidean inner product. OLS can thus be interpreted as a weighted averaging procedure, where the predicted outcome \( \hat{y}_j \) is a sum of observed responses \( y_i \), weighted by \( \omega_{ji} \), which reflects the similarity between the corresponding factor \( F_i \) and \( F_j \) in the Euclidean sense. \cite{dual} discuss this property as a way to interpret the weights \( \omega_{ji} \) in high-dimensional and nonlinear machine learning-based macroeconomic forecasts. 

Equation~\eqref{innerprod} factorizes into scale and alignment via $\omega_{ji} = \|F_j\|\|F_i\|\cos(\theta_{ji})$, so the largest weights arise when the encoded test and training vectors are both large in norm and well aligned in direction. Representations similar to \eqref{gram} appear in textbooks for fitted values (\citet{hastie2009elements}, on the SVD of the hat matrix); the novelty here is reinterpreting them as a similarity-based bridge to attention. The factor structure arising from the OLS fit is \textit{entirely mechanical}: a purely algebraic reorganization that holds for any dataset, regardless of distribution, stationarity, or sample size.

\looseness=-1 This invites a comparison with $k$-nearest neighbors. $k$-NN puts equal weight on a few neighbors and none elsewhere. OLS gives every training point a weight that varies smoothly with where the test point sits, and that weight can be negative: an observation that is the polar opposite of the test point in the orthonormal space contributes with the opposite sign, which no nearest-neighbor rule can do. Softmax attention sits in between, accentuating strong similarities and attenuating weak ones.

\paragraph{On the Choice of the Spectral Decomposition.} While there are multiple ways to decompose a symmetric positive definite matrix, the spectral decomposition is particularly useful for understanding OLS as a similarity-matching system in an uncorrelated \( P \)-dimensional space. When characteristics are correlated, aggregating similarity scores across dimensions is not straightforward. Using 
\begin{align}\label{innerprod2}
\hat{y}_{j} &= \sum_{i=1}^{N} \langle X_{j}, X_i \rangle y_i
\end{align} 
instead of equation \eqref{innerprod}, even when $\boldsymbol{X}_{\text{train}}' \boldsymbol{X}_{\text{train}} \neq I_P$, is suboptimal because it fails to account for covariances between predictors in computing similarities. For example, if two predictors are highly correlated with $|\rho| \approx 1$, equation \eqref{innerprod2} effectively counts the redundant information twice, disproportionally affecting the estimated prediction weights.

However, in an orthonormal basis where \textit{characteristics} are mutually uncorrelated, the total similarity between observations $i$ and $j$ can be computed simply as the sum of similarities for each characteristic calculated separately, as in equation \eqref{innerprod}. This is analogous to standard results on variance calculations: when \textit{observations} are mutually uncorrelated, the variance of a sum equals the sum of individual variances. By the same logic, the total similarity in this transformed orthonormal space is just the sum of the similarities of individual characteristics. When $\boldsymbol{X}_{\text{train}}' \boldsymbol{X}_{\text{train}}$ is singular ($P > N$), the same reading holds with a regularized inverse such as ridge, on the subspace the data span. The spectral choice is a normalization: any factorization $(\boldsymbol{X}_{\text{train}}'\boldsymbol{X}_{\text{train}})^{-1}=\boldsymbol{M}\boldsymbol{M}'$ gives the same inner products and predictions, so the embedding is identified only up to a rotation, as factors are in a factor model.

\paragraph{Encoder/decoder view.}
More generally, we can define two mappings,
\begin{align}
\boldsymbol{W}_{\text{train}}: \mathbb{R}^{P} \to \mathbb{R}^{P} \quad \quad \text{and} \quad \quad   \boldsymbol{W}_{\text{test}}: \mathbb{R}^{P} \to \mathbb{R}^{P},
\end{align}
which embed $\boldsymbol{X}_{\text{train}}$ and $\boldsymbol{X}_{\text{test}}$ into a space where their similarity can be directly measured via Euclidean inner products. In the case of the OLS solution, these mappings take the form
\begin{align}\label{ols_wmap}
{\boldsymbol{W}}_{\text{train}} = {\boldsymbol{W}}_{\text{test}} = \boldsymbol{U} \Lambda^{-\frac{1}{2}},
\quad
\boldsymbol{F}_{\text{train}} = \boldsymbol{X}_{\text{train}} \boldsymbol{W}, \quad
\boldsymbol{F}_{\text{test}} = \boldsymbol{X}_{\text{test}} \boldsymbol{W},
\end{align}
\looseness=-1 so that the OLS prediction takes the inner-product form $\hat{\boldsymbol{y}}_{\text{test}} = \boldsymbol{F}_{\text{test}} \boldsymbol{F}_{\text{train}}'\boldsymbol{y}$ stated in~\eqref{gram}. These transformations define what are often referred to as encoding and decoding operations. The encoding step transforms the original predictors into the factor space, where relationships among predictors are orthogonalized for easier comparison. The decoding step ensures that the \emph{same} transformation applied to the training set is also applied to the test set, preserving structural alignment and relying solely on mappings parameterized by in-sample data. The encoding is fairly passive from a modeling standpoint, since it is neither compressing nor expanding the information available in $\boldsymbol{X}$ ($\boldsymbol{W}\colon \mathbb{R}^P\to\mathbb{R}^P$). Still, despite lacking a factorization pre-processing step or an explicit encoding layer, OLS inherently performs orthogonal encoding and decoding operations in the background. Section~\ref{sec:reducdim} will let $\boldsymbol{W}$ project into lower-dimensional subspaces, further aligning the framework with factor models and even autoencoders.

\subsection{The Attention Mechanism}

\looseness=-1 The standard attention mechanism \citep{vaswani2017attention} is defined as
\begin{align}
  \text{Attention}(\mathbf{Q}, \mathbf{K}, \mathbf{V}) = \text{softmax} \left( \frac{\mathbf{Q} \mathbf{K}'}{\sqrt{P}} \right) \mathbf{V},
\end{align}
where \( \mathbf{Q} \) (queries), \( \mathbf{K} \) (keys), and \( \mathbf{V} \) (values) are linear projections of input data, and $\text{softmax}(z_i) = e^{z_i}/\sum_{\tau} e^{z_\tau}$ normalizes attention weights. The mechanism can be understood as similarity-based retrieval: queries search for relevant keys, and the corresponding values are aggregated with soft weights. This connects to kernel methods, where attention weights correspond to similarity measures in a latent space \citep{tsai2019transformer, choromanski2020rethinking}:
\begin{align}\label{kernelview}
  \text{Attention}(\mathbf{Q}, \mathbf{K}, \mathbf{V}) = \boldsymbol{\gamma} \mathbf{V}, \qquad \gamma_{i\tau} = \frac{k(\mathbf{q}_i, \mathbf{k}_\tau)}{\sum_{s=1}^N k(\mathbf{q}_i, \mathbf{k}_s)} \quad \text{for a kernel } k(\cdot, \cdot).
\end{align}

\subsection{The OLS-Attention Nexus}\label{sec:nexus}

Using the encoder-decoder mappings of \eqref{ols_wmap}, the OLS prediction is rewritten in an attention-like form via $\mathbf{Q}$, $\mathbf{K}$, $\mathbf{V}$:
\begin{align*}
  \mathbf{Q} &= \boldsymbol{X}_{\text{test}} \boldsymbol{W}_Q , \\
  \mathbf{K} &= \boldsymbol{X}_{\text{train}} \boldsymbol{W}_K, \\
  \mathbf{V} &= \boldsymbol{y} \cdot \boldsymbol{w}_v.
\end{align*}
To generalize this formulation, I replace the softmax function with a general function \( g(\cdot) \) and substitute the expressions for \( \mathbf{Q} \), \( \mathbf{K} \), and \( \mathbf{V} \):
\begin{align}
  \text{Attention}(\boldsymbol{X}_{\text{test}}, \boldsymbol{X}_{\text{train}}, \mathbf{y}; \mathbf{W}) = g \left( \boldsymbol{X}_{\text{test}} \boldsymbol{W}_Q \boldsymbol{W}_K' \boldsymbol{X}_{\text{train}}' \right) \boldsymbol{y} \cdot \boldsymbol{w}_v.
\end{align}
Now, setting \( g(\cdot) = I(\cdot) \) (the identity function), \( \boldsymbol{W}_Q \boldsymbol{W}_K' = (\boldsymbol{X}_{\text{train}}'\boldsymbol{X}_{\text{train}})^{-1} \) or equivalently $\boldsymbol{W}_K=\boldsymbol{W}_Q= \boldsymbol{U} \Lambda^{-\frac{1}{2}}$ as in equation \eqref{ols_wmap}, and \( \boldsymbol{w}_v = \mathbf{1} \), we obtain:
\begin{align}\label{ols:att}
  \text{Attention}(\boldsymbol{X}_{\text{test}}, \boldsymbol{X}_{\text{train}}, \mathbf{y}; \mathbf{W}) &= g \left( \boldsymbol{X}_{\text{test}} \boldsymbol{W}_Q \boldsymbol{W}_K' \boldsymbol{X}_{\text{train}}' \right) \boldsymbol{y} \cdot \boldsymbol{w}_v \\
  &= \boldsymbol{X}_{\text{test}} (\boldsymbol{X}_{\text{train}}' \boldsymbol{X}_{\text{train}})^{-1} \boldsymbol{X}_{\text{train}}' \boldsymbol{y} \\
  &= \boldsymbol{X}_{\text{test}} \hat{\boldsymbol{\beta}}.
\end{align}
Thus, the out-of-sample OLS prediction can be interpreted as a form of scaled attention, where each test sample attends to the training observations through a similarity measure encoded in an orthonormal \( P \)-dimensional space. The expression $\boldsymbol{W}_Q \boldsymbol{W}_K' = (\boldsymbol{X}_{\text{train}}' \boldsymbol{X}_{\text{train}})^{-1}$ is derived directly from the least squares closed-form solution as one embedding \textit{that minimizes squared prediction error}.

In information-retrieval language: the query $\mathbf{Q}$ encodes $\boldsymbol{X}_{\text{test}}$, the keys $\mathbf{K}$ encode $\boldsymbol{X}_{\text{train}}$ via the same map, and the values $\mathbf{V}$ are the realized outcomes $\boldsymbol{y}$, linearly combined to produce a prediction. The setup as written is \textit{cross-attention} (queries from a different input than keys); replacing $\mathbf{Q} = \boldsymbol{X}_{\text{train}}\boldsymbol{W}_Q$ recovers in-sample fitted values and is closer to \textit{self-attention}, treated in Appendix~\ref{VARs}. Section~\ref{sec:nl} reintroduces a nonlinear $g(\cdot)$ while preserving the regression interpretation.

\paragraph{The OLS Solution as an Optimal Embedding.}
The standard interpretation of least squares solves for the optimal coefficient vector. The alternative formulation optimizes the embedding transformation itself: rather than minimizing over $\boldsymbol{\beta}$, it minimizes over $\boldsymbol{\Omega}$ the squared error of the similarity rule, and the two problems deliver the same predictions. The plug-in choice $\boldsymbol{W}_Q\boldsymbol{W}_K' = (\boldsymbol{X}_{\text{train}}'\boldsymbol{X}_{\text{train}})^{-1}$ is not arbitrary: it is a closed-form minimizer of the embedding problem
\begin{align}\label{ols_w}
  \hat{\boldsymbol{\Omega}} \;=\; (\boldsymbol{X}_{\text{train}}' \boldsymbol{X}_{\text{train}})^{-1} \;\in\; \arg\min_{\boldsymbol{\Omega} \in \mathbb{R}^{P\times P}} \| \boldsymbol{y} - \boldsymbol{X}_{\text{train}} \boldsymbol{\Omega} \boldsymbol{X}_{\text{train}}' \boldsymbol{y} \|^2,
\end{align}
where $\boldsymbol{\Omega}\equiv\boldsymbol{W}_Q\boldsymbol{W}_K'$. Is $(\boldsymbol{X}_{\text{train}}'\boldsymbol{X}_{\text{train}})^{-1}$ the only embedding that attains the minimum? Not for one realized target. The objective depends on $\boldsymbol{\Omega}$ only through the vector $\boldsymbol{\Omega}\boldsymbol{X}_{\text{train}}'\boldsymbol{y}$, so any $\boldsymbol{\Omega}$ that agrees with $(\boldsymbol{X}_{\text{train}}'\boldsymbol{X}_{\text{train}})^{-1}$ on that single vector gives the same fitted values and the same minimum, and some of those $\boldsymbol{\Omega}$ have rank one. An embedding is a geometry of the predictor space, and it should not depend on which outcome is regressed on $\boldsymbol{X}$. So the requirement is that one $\boldsymbol{\Omega}$ minimize the objective for every outcome vector, the realized one and all others. Under that requirement the minimizer is unique, and it is $(\boldsymbol{X}_{\text{train}}'\boldsymbol{X}_{\text{train}})^{-1}$ (Appendix~\ref{app:unique}).

Estimating $\boldsymbol{\Omega}$ directly is not computationally advantageous in the linear case---$P(P+1)/2$ parameters instead of $P$ coefficients---but the formulation is what makes the bridge to attention precise: in a linear attention module whose $\boldsymbol{W}_Q$ and $\boldsymbol{W}_K$ are learned by backpropagation, any squared-error optimum delivers the OLS prediction. From the regression view, $\boldsymbol{W}_Q\boldsymbol{W}_K'$ plays the role of a precision matrix: the inverse covariance of the variables entering the module, generated by the layers below in a Transformer and equal to $(\boldsymbol{X}_{\text{train}}'\boldsymbol{X}_{\text{train}})^{-1}$ here. A precision matrix is symmetric and positive definite, so it factors as one map applied to queries and keys alike, as in equation~\eqref{ols_wmap}, and the view suggests that little is lost by setting $\boldsymbol{W}_K=\boldsymbol{W}_Q$. Tying the two has been tried, mostly at little cost \citep{tsai2019transformer,kitaev2020reformer}, with a modest loss reported for causal language modeling \citep{kayyam2026three}.

\vspace{0.75em}
\section{Beyond OLS}\label{sec25} %

The equivalence between the attention module and OLS relies on several simplifications with respect to what is actually going on under the hood of Transformers. This section adds them back one at a time, starting from OLS. I begin by discussing modifications to the attention module itself, followed by its integration within the broader Transformer architecture.

\subsection{Dimensionality Reduction and Shrinkage}\label{sec:reducdim}

The OLS solution reveals that the encoding and decoding matrices, \( \boldsymbol{W}_K \) and \( \boldsymbol{W}_Q \), serve solely to project \( \boldsymbol{X} \) into a space where features are orthogonal, facilitating direct computation of Euclidean inner products. In Transformer-based models, by contrast, the learned mappings \( \boldsymbol{W}_K \) and \( \boldsymbol{W}_Q \) project the $d$-dimensional token representations into a lower-dimensional space, $d/h$ per head, while the embedding at the entrance and the unembedding at the exit move between the vocabulary and that $d$-dimensional space.

How does the OLS-based attention mechanism compare to the learned embeddings in Transformers? The answer lies in Principal Component Regression (PCR). Suppose we redefine 
\begin{align}
\boldsymbol{W}_{\text{train}}: \mathbb{R}^{P} \to \mathbb{R}^{L} \quad \quad \text{and} \quad \quad  \boldsymbol{W}_{\text{test}}: \mathbb{R}^{P} \to \mathbb{R}^{L},
\end{align}
as mappings from a space of dimension \( P \) to a lower-dimensional space \( L \), where \( L < P \), effectively compressing the data representation. In this case, the resulting formulation is closely related to the application of PCA to reduce the dimensionality of the predictor space before applying OLS.

Specifically, consider truncating the eigenvalue decomposition of the sample covariance matrix:
\begin{align}
  \boldsymbol{X}_{\text{train}}' \boldsymbol{X}_{\text{train}} = \boldsymbol{U} \Lambda \boldsymbol{U}',
\end{align}
where, instead of inverting \( \Lambda \) directly, we define a low-rank approximation by retaining only the first \( L \) principal components, setting the smallest eigenvalues to zero. This corresponds to constructing \( \boldsymbol{W}_K \) and \( \boldsymbol{W}_Q \) such that they select the leading eigenvectors: 
\[
\boldsymbol{W}_K = \boldsymbol{W}_Q = \boldsymbol{U}_L \Lambda_L^{-\frac{1}{2}},
\]
where \( \Lambda_L \in \mathbb{R}^{L \times L} \) is a diagonal matrix containing the first \( L \) eigenvalues, and \( \boldsymbol{U}_L \in \mathbb{R}^{P \times L} \) holds the corresponding \( L \) eigenvectors. Therefore, we have $ \boldsymbol{F}_{\text{test}}^{L} = \boldsymbol{X}_{\text{test}} \boldsymbol{W}_Q $ and $\boldsymbol{F}_{\text{train}}^{L} = \boldsymbol{X}_{\text{train}} \boldsymbol{W}_K $, both being lower-dimensional representations of the original data projected onto the subspace spanned by the first $L$ principal components.

In this formulation, it is straightforward to see that PCR's predictions--essentially performing OLS on \(\boldsymbol{F}_{\text{train}}^L\), the top \(L\) principal components of the input data--reduce to our linear attention mechanism with dimensionality reduction:
\begin{align}\label{pcr_pred}
\hat{\boldsymbol{y}}_{\text{test}} = \boldsymbol{F}_{\text{test}}^{L} \hat{\boldsymbol{\theta}} 
= \boldsymbol{F}_{\text{test}}^{L} 
\underbrace{\left(\boldsymbol{F}_{\text{train}}^{L'} \boldsymbol{F}_{\text{train}}^{L}\right)^{-1}}_{= \boldsymbol{I}_L} 
\boldsymbol{F}_{\text{train}}^{L'} \boldsymbol{y}_{\text{train}}
= \boldsymbol{F}_{\text{test}}^{L} \boldsymbol{F}_{\text{train}}^{L'} \boldsymbol{y}_{\text{train}}.
\end{align}
The embedding behind this rule is $\boldsymbol{W}_Q\boldsymbol{W}_K'=\boldsymbol{U}_L \Lambda_L^{-1} \boldsymbol{U}_L'$, the least-squares embedding of Section~\ref{sec2} with its last $P-L$ directions removed. Written as an embedding, \eqref{pcr_pred} reads
\begin{align}\label{pcr_embed}
\hat{\boldsymbol{y}}^{\text{PCR}}_{\text{test}}=\boldsymbol{F}^{L}_{\text{test}}\boldsymbol{F}^{L\prime}_{\text{train}}\boldsymbol{y}
= \boldsymbol{X}_{\text{test}}\,\underbrace{\boldsymbol{U}_L \Lambda_L^{-1} \boldsymbol{U}_L'}_{\hat{\boldsymbol{\Omega}}_L}\,\boldsymbol{X}_{\text{train}}'\boldsymbol{y}
\qquad\text{for every } \boldsymbol{y}.
\end{align}
Is this embedding optimal in the sense of Section~\ref{sec2}? Add the rank constraint to the embedding problem, imposed on the data side: the embedding is confined to the leading $L$ principal directions of $\boldsymbol{X}_{\text{train}}'\boldsymbol{X}_{\text{train}}$, $\boldsymbol{\Omega}=\boldsymbol{U}_L\boldsymbol{D}\boldsymbol{U}_L'$ with $\boldsymbol{D}\in\mathbb{R}^{L\times L}$, the Eckart--Young truncation of the covariance rather than of its inverse,
\begin{align}
  \min_{\boldsymbol{\Omega}} \| \boldsymbol{y} - \boldsymbol{X}_{\text{train}} \boldsymbol{\Omega} \boldsymbol{X}_{\text{train}}' \boldsymbol{y} \|^2 \quad \text{s.t.}\ \ \text{rank}(\boldsymbol{\Omega})=L,\ \boldsymbol{\Omega}=\boldsymbol{U}_L\boldsymbol{D}\boldsymbol{U}_L', \qquad \text{for every } \boldsymbol{y}.
\end{align}
The unique solution is $\boldsymbol{D}=\Lambda_L^{-1}$, that is, $\hat{\boldsymbol{\Omega}}_L$ (Appendix~\ref{app:unique}). The rank constraint alone would not do: for one realized $\boldsymbol{y}$ a rank-one embedding already attains the OLS fit, and no rank-deficient embedding serves every $\boldsymbol{y}$. PCR keeps the largest-variance directions and drops the rest, which is dimension reduction in the usual sense.

\paragraph{Ridge, and Three Estimators in One Module.}
Ridge regression fits the same template. Its prediction is $\boldsymbol{X}_{\text{test}}\boldsymbol{\Omega}_\lambda\boldsymbol{X}_{\text{train}}'\boldsymbol{y}$ with $\boldsymbol{\Omega}_\lambda=(\boldsymbol{X}_{\text{train}}'\boldsymbol{X}_{\text{train}}+\lambda\boldsymbol{I})^{-1}=\boldsymbol{U}(\Lambda+\lambda\boldsymbol{I})^{-1}\boldsymbol{U}'$, so the ridge embedding keeps the rotation $\boldsymbol{U}$ of least squares and only changes the scaling of each direction, from $1/\lambda_i$ to $1/(\lambda_i+\lambda)$. Shrinkage acts on the embedding, by damping the low-variance directions, rather than on the coefficients.

Together with \eqref{pcr_embed}, the OLS and ridge embeddings give three estimators that differ only in the diagonal of one embedding, $\boldsymbol{\Omega}=\boldsymbol{U}\boldsymbol{D}\boldsymbol{U}'$ with
\begin{align}\label{three_embed}
\boldsymbol{D}_{\text{OLS}}=\Lambda^{-1},\qquad
\boldsymbol{D}_{\text{ridge}}=(\Lambda+\lambda\boldsymbol{I})^{-1},\qquad
\boldsymbol{D}_{\text{PCR}}=\text{diag}\bigl(\lambda_1^{-1},\dots,\lambda_L^{-1},0,\dots,0\bigr),
\end{align}
inside the same attention module. These embeddings are used below: the ridge embedding initializes every head of the Attention Regression experiments (Appendix~\ref{sec:simulation}) and warm-starts the Regression Block (Appendix~\ref{app:regblk}).

\paragraph{Beyond PCR: Covariance-Estimation Routes to $\boldsymbol{\Omega}$.}
\looseness=-1 Reading the optimal embedding as $\hat{\boldsymbol{\Omega}} = (\boldsymbol{X}'\boldsymbol{X})^{-1}$ opens a direct pipeline from the precision-matrix literature to attention. Whenever $P$ is large relative to $N$, the sample inverse covariance is unstable, and one would shrink, sparsify, or factor it---each operation has a precise counterpart on $\boldsymbol{W}_Q\boldsymbol{W}_K'$. Linear shrinkage \citep{ledoit2004wellconditioned} corresponds to ridge-regularized attention; thresholding \citep{bickel2008covariance} to attention on a sparsified covariance; factor-based estimation \citep{fan2013large} to PCR-style attention with an idiosyncratic diagonal.

\subsection{Nonlinearity and the Use of Softmax}\label{sec:nl}

Reintroducing nonlinearity—along with the dimension reduction discussed above—leads us to the full attention mechanism used in Transformer architectures. The row-wise application of the softmax function \( g(\cdot) \) brings nonlinearities into the attention-based regression framework, making it impossible to derive \( \boldsymbol{\Omega} \) in closed form from
\begin{align}\label{att:reg}
  \min_{\boldsymbol{\Omega} \in \mathbb{R}^{P \times P}} \| \boldsymbol{y} - \text{softmax} \left( \boldsymbol{X}_{\text{train}} \boldsymbol{\Omega} \boldsymbol{X}_{\text{train}}' \right) \boldsymbol{y} \|^2 \, .
\end{align}
The softmax produces strictly positive attention weights summing to one, so predictions are non-negative convex combinations of training outcomes. In short: there is no such thing as \textit{negative} attention---while OLS makes use of it all the time. This rules out the signed, symmetry-exploiting weights OLS happily uses and makes the resulting estimator a row-stochastic close cousin of Random Forest \citep{dual}.

Row-wise normalization is not what makes \eqref{att:reg} new: provided $\boldsymbol{X}_{\text{train}}$ contains an intercept, the rows of the OLS projection matrix already sum to one. The genuine nonlinear ingredient is the exponential transformation, which squashes negatives and amplifies large positives. Equation~\eqref{att:reg} can thus be read as a flexible nonparametric regression with a softmax-induced simplex constraint on proximity weights---a hybrid sitting between kernel methods (smooth nonlinearity) and tree-based methods (non-negative weights). This is what I call the \emph{Attention Regression}, and Section~\ref{sec:exp} puts it to work.

\subsection{Stacking Layers}\label{sec:depth}
The identity is a one-layer statement, so three things need to be checked in turn: several linear attention layers acting in sequence, the same with residual connections, and nonlinearity between layers. The short answer is that the first two change nothing, and the third turns the claim from a global one into a local one.

Let the \emph{linear} layers act in sequence, each on the output of the previous one, with queries and keys formed from that output, as in a real network. Each layer multiplies its input by a fixed matrix, so the output of the whole stack is still the original $\boldsymbol{X}$ times a fixed matrix: a stack of linear attention layers is one linear attention layer on the original inputs. In the least-squares case this is the familiar fact that regressing fitted values on $\boldsymbol{X}$ returns the same fitted values. Residual connections change the algebra but not the conclusion. Each residual block also multiplies its input by a fixed matrix, generically invertible, so the whole network followed by a linear output layer is a factorization of one linear predictor $\boldsymbol{X}\boldsymbol{\beta}_{\text{eff}}$, and since $\boldsymbol{\beta}_{\text{eff}}$ is unrestricted its squared-error optimum is exactly the OLS prediction. The statement concerns the optimum, whatever path optimization takes to reach it. Linear depth reparametrizes the estimator without enlarging its function class. The algebra is in Appendix~\ref{app:stack}.

Nonlinear transformations change the situation because they alter the representation between attention operations. A standard Transformer does this principally through its feed-forward network, and LayerNorm also makes subsequent operations depend on a transformed representation. Once these components intervene, later attention modules no longer act on the original $\boldsymbol{X}$, and the exact collapse no longer holds end to end. The claim is therefore local: conditional on the representation entering a restricted attention module, the regression interpretation applies to that module. Suppose the nonlinear map is represented, or locally approximated, by polynomial functions of a small number of learned regression directions. The model is then a structured low-rank polynomial regression, with interactions flowing only through those few directions (Appendix~\ref{app:stack}). The Regression Block of Section~\ref{sec:exp} is built in the spirit of this principle: polynomial features, a compact PCA basis, then a supervised combination of the retained components.

\section{Empirical Evidence}\label{sec:exp}

The equivalence developed above raises two empirical questions. Equation~\eqref{att:reg} is a flexible nonparametric estimator with a learned similarity metric, so it is reasonable to ask how it fares against off-the-shelf nonlinear methods. Going the other way, if the predictive content of an attention block is essentially regression-like, then replacing the softmax-attention sublayer of a transformer with explicit regression on expanded features should not cost much, and might gain something. I address the first question on simulated and real data, and the second on real data alone.

\begin{figure}[t]
 \centering
 \includegraphics[width=0.95\linewidth]{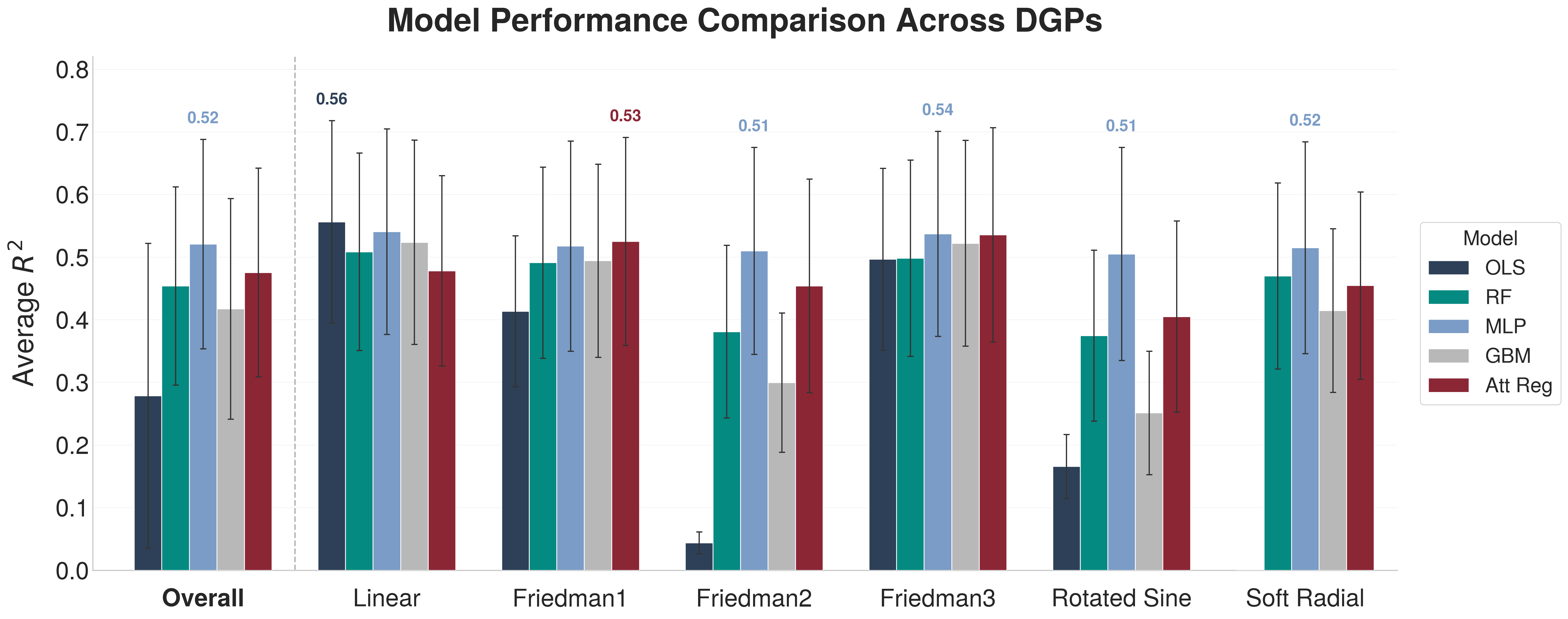}
 \caption{Monte Carlo: average out-of-sample $R^2$ across $N\in\{500,1000,2500,5000\}$ and SNR$\in\{0.5,1,2,3\}$ for each DGP. Error bars are $\pm 1$ standard deviation across the 16 conditions per DGP. Full per-condition table in Appendix~\ref{sec:simulation}.}
 \label{fig:model_performance}
\end{figure}

\paragraph{Simulated data.}
The Monte Carlo experiment uses six standard regression DGPs (Linear, Friedman~1--3, Rotated Sine, Soft Radial), four sample sizes $N\in\{500,1000,2500,5000\}$, and four signal-to-noise ratios SNR$\in\{0.5,1,2,3\}$, with results averaged over 10 replications. Attention Regression (five heads, no per-DGP tuning) is compared against OLS, Random Forest, Gradient Boosting, and an MLP; full setup and the per-DGP table are in Appendix~\ref{sec:simulation}. Across the 96 conditions, Attention Regression averages $R^2=0.48$, behind the MLP ($0.52$), close to Random Forest ($0.45$), and ahead of Gradient Boosting ($0.42$) and OLS ($0.28$). It does particularly well on the interaction-heavy Friedman designs and on Rotated Sine. Figure~\ref{fig:model_performance} reports the per-DGP picture.

\paragraph{Real data.}
Eight standard UCI/OpenML tabular benchmarks (California, Yacht, Energy, Concrete, Airfoil, Abalone, Kin8nm, Protein) provide the test bed \citep{grinsztajn2022why,gorishniy2021revisiting}, with 5 random 80/20 splits, one fixed configuration per method, features and targets standardized on the training split, and a cap of $N=5000$ on the larger datasets (lifted in Appendix~\ref{app:scope}). The baselines were chosen for three reasons. Least squares is the exact linear case of the equivalence, Random Forest and an MLP are standard strong conventional baselines, and the FT-Transformer \citep{gorishniy2021revisiting} is the canonical attention architecture for tabular data. Pretrained tabular foundation models answer a different question and appear as reference points in Appendix~\ref{app:scope}. They lead, as production-grade systems should, and the distance of a Regression Block trained from scratch with 30K parameters is modest.

\noindent\textbf{Introducing the Regression Block.} It is a drop-in replacement for the multi-head softmax attention sublayer, inside a single block that keeps the FT-Transformer's residual-plus-LayerNorm pattern and its position-wise FFN. In place of $\text{softmax}(\boldsymbol{Q}\boldsymbol{K}'/\sqrt{d})\boldsymbol{V}$, the token representations are mapped to degree-2 polynomial cross-features, compressed by a frozen PCA to 200 components, and passed through a linear regression layer initialized from a ridge fit. There is no kernel and no learned similarity metric, only regression coefficients. The Regression Block is one layer deep against FT-Transformer's three and totals about $30$K trainable parameters against $100$K (full architecture, and the differences from the FT-Transformer skeleton, in Appendix~\ref{app:regblk}). The Regression Block is the most literal test of the title: a working transformer architecture in which OLS-style regression has taken the place of attention.

\begin{table}[t]
\centering

\caption{Out-of-sample $R^2$ on UCI/OpenML tabular regression benchmarks. }
\label{tab:realdata_main}
\footnotesize
\setlength{\tabcolsep}{4.5pt}
\renewcommand{\arraystretch}{1.0}
\begin{tabular}{@{}l r r cccc cc@{}}
\toprule
& & & \multicolumn{4}{c}{\textit{Baselines}} & \multicolumn{2}{c}{\textit{Contributions}} \\
\cmidrule(lr){4-7}\cmidrule(lr){8-9}
\textbf{Dataset} & \boldmath$N$ & \boldmath$P$ & \textbf{OLS} & \textbf{RF} & \textbf{MLP} & \textbf{FT-T} & \textbf{Att.~Reg} & \textbf{Reg.~Block} \\
\midrule
California & 5000 & 8 & 0.597 & \textbf{\color{ForestGreen}0.777} & 0.763 & 0.764 & 0.738 & \textbf{0.766} \\
Yacht   & 308 & 6 & 0.562 & 0.979 & 0.970 & \textbf{0.988} & 0.958 & \textbf{\color{ForestGreen}0.989} \\
Energy   & 768 & 8 & 0.913 & \textbf{\color{ForestGreen}0.996} & 0.994 & 0.994 & \textbf{0.995} & \textbf{0.995} \\
Concrete  & 1030 & 8 & 0.624 & 0.892 & 0.895 & \textbf{0.898} & 0.837 & \textbf{\color{ForestGreen}0.907} \\
Airfoil  & 1503 & 5 & 0.497 & 0.910 & 0.917 & \textbf{0.924} & 0.757 & \textbf{\color{ForestGreen}0.927} \\
Abalone  & 4177 & 7 & 0.260 & 0.322 & 0.323 & \textbf{0.324} & 0.312 & \textbf{\color{ForestGreen}0.329} \\
Kin8nm   & 5000 & 8 & 0.430 & 0.664 & 0.912 & \textbf{0.914} & 0.851 & \textbf{\color{ForestGreen}0.920} \\
Protein  & 5000 & 9 & 0.285 & \textbf{\color{ForestGreen}0.521} & \textbf{0.486} & 0.426 & 0.361 & 0.455 \\
\bottomrule
\end{tabular}

\vspace{0.1em}
\begin{minipage}{0.97\linewidth}
\footnotesize
\textit{Notes:} Out-of-sample $R^2$ averaged over 5 random 80/20 splits; larger datasets capped at $N=5000$. Att.~Reg is Eq.~\eqref{att:reg}, Reg.~Block the Regression Block of Appendix~\ref{app:regblk}. \textbf{\color{ForestGreen}Green bold} = best, \textbf{black bold} = second best.
\end{minipage}
\end{table}

Table~\ref{tab:realdata_main} reports out-of-sample $R^2$ and answers the two questions in turn. On the first, Attention Regression, with five heads and no per-dataset tuning, lands in the neighborhood of Random Forest and the MLP, which is enough for a proof of concept. On the second, replacing the attention sublayer by an explicit regression does not hurt: the Regression Block is ahead of the FT-Transformer on all eight datasets, at roughly thirty percent of the parameters and a third of the depth. Six of the eight margins are within seed noise. The evidence is the consistency of the sign across the eight datasets and the pooled paired difference (Appendix~\ref{app:protocol}). The gap between OLS and Random Forest, wide on six of the eight datasets, says there is nonlinearity to capture, and the ablations below say the polynomial expansion is what captures it.

\paragraph{Ablation Studies.} Appendix~\ref{app:ablations} takes the Regression Block apart in ten configurations, all run under the protocol of Table~\ref{tab:realdata_main}. The polynomial expansion is by far the most important ingredient, and its payoff tracks the paper's own nonlinearity diagnostic: across datasets, the loss from removing it rises with the gap between OLS and Random Forest. The PCA compression buys parameter economy rather than accuracy, since removing it costs little at nearly six times the parameters. Capacity does not explain the comparison either way: neither shrinking the FT-Transformer to the Regression Block's budget nor widening the Regression Block to the FT-Transformer's changes it, and stacking the Regression Block to three blocks makes it worse. The explicit regression carries the result, since softmax attention in the same skeleton is far behind.

\paragraph{Additional Experiments and Observations.} The same Regression Block, unchanged and untuned, was then tried beyond Table~\ref{tab:realdata_main} (Appendices~\ref{app:ablations} and~\ref{app:scope}). First, without the sample-size cap the two architectures stay close, the Regression Block ahead on six of eight. Second, on inputs of up to 81 dimensions it stays within $0.026$ of the FT-Transformer and is ahead on the highest-dimensional dataset. It can be cut to about 7,500 parameters, 7.4 percent of the FT-Transformer's, and stay within $0.005$ of it in mean $R^2$. Third, with a cross-entropy head in place of the squared-error one, it carries over to classification, with the best average rank and mean accuracy among the methods trained from scratch on six datasets.

\section{Closing Remarks and Limitations}\label{sec3}

This paper shows that ordinary least squares can be reformulated as a constrained attention mechanism, revealing a formal connection between classical regression and the core structure underlying attention-based models. By interpreting OLS as a similarity-based method in an orthonormal space, we recover key features associated with attention---dimensionality reduction, learned representations, and projection-based prediction---from standard regression techniques. Seen from the regression side, attention is nonlinear least squares with generated features, and contextual embeddings are its fitted values.

\paragraph{Implications.} Attention is not as distant from the traditional statistical toolkit as it might initially appear, and hopefully this understanding can prove useful to both lines of work. One concrete example: the closed form $(\boldsymbol{X}'\boldsymbol{X})^{-1}$ replaces backpropagation in the linear case and warm-starts it in the nonlinear one, as in Section~\ref{sec:exp}. Another is a design question: least squares uses negative attention all the time, so the simplex constraint of softmax is a choice, and Section~\ref{sec:exp} shows what giving it up buys on tabular data.

\paragraph{Limitations.} The exact identity covers one restricted attention module, and in a full Transformer the reading is local to each module (Section~\ref{sec:depth}). On the empirical side, the Regression Block's cross-features grow quadratically with the number of inputs, which becomes the binding constraint beyond the dimensions of Table~\ref{tab:realdata_main} (Appendix~\ref{app:scope}).

\begin{ack}
The experiments of Appendices~\ref{app:protocol}--\ref{app:scope} ran on Modal under a Modal for Academics compute grant; the author received no other third-party funding and declares no competing interests.
\end{ack}

\clearpage

\setlength\bibsep{5pt}

\bibliographystyle{plainnat}
\bibliography{references}

\newpage
\appendix
\raggedbottom  %

\section{Further Results on the Equivalence}\label{sec:tr}

This section first proves the uniqueness result of Section~\ref{sec2}, then outlines key components, including multi-head and self-attention, pre-trained word embeddings, time series features such as masking, pooled (or global) panel estimation, and the recursive chaining of multiple attention modules.

The discussion in the main text has focused on the attention mechanism itself, treating it as a distinct entity. However, its role within Transformer architectures involves numerous operations both before data reaches the attention modules and after it passes through them. The original Transformer architecture by \cite{vaswani2017attention} consists of 6 encoder and 6 decoder layers, each combining multi-head self-attention with feedforward networks, wrapped in residual connections and layer normalization. At each layer, the attention output is added to the input and passed through a feedforward network, with this process recursively repeated across layers.

\subsection{Uniqueness of the Optimal Embedding}\label{app:unique}

\paragraph{One realized target.}
For a single realized target $\boldsymbol{y}$ the objective does not identify the embedding. Write $\boldsymbol{S}=\boldsymbol{X}_{\text{train}}'\boldsymbol{X}_{\text{train}}$ and $\boldsymbol{z}=\boldsymbol{X}_{\text{train}}'\boldsymbol{y}$. Minimizing \eqref{ols_w} requires only $\boldsymbol{\Omega}\boldsymbol{z}=\boldsymbol{S}^{-1}\boldsymbol{z}$, so the solution set is the affine set
\[
\boldsymbol{\Omega}=\boldsymbol{S}^{-1}+\boldsymbol{\Delta},\qquad \boldsymbol{\Delta}\boldsymbol{z}=\boldsymbol{0},
\]
of dimension $P(P-1)$. The prediction identity of Section~\ref{sec2} is unaffected, every member of that set returning the same fitted values for the target that defined it.

\paragraph{Every target.}
Uniqueness holds once the embedding is required to minimize the objective for every outcome vector, not only for the one realized. With $\text{rank}(\boldsymbol{X}_{\text{train}})=P$, OLS uniquely minimizes squared error at each $\boldsymbol{y}$, with $\hat{\boldsymbol{\beta}}(\boldsymbol{y})=\boldsymbol{S}^{-1}\boldsymbol{X}_{\text{train}}'\boldsymbol{y}$, so
\[
\boldsymbol{\Omega}\boldsymbol{z}=\boldsymbol{S}^{-1}\boldsymbol{z}\quad\text{for every }\boldsymbol{z}\text{ in the range of }\boldsymbol{X}_{\text{train}}',
\]
which is all of $\mathbb{R}^P$. Hence $\boldsymbol{\Omega}=\boldsymbol{S}^{-1}$. The full-rank condition does two standard things: it makes $\hat{\boldsymbol{\beta}}$ unique at each target and it makes the targets sweep all of $\mathbb{R}^P$. The inverse Gram matrix is not determined by one realized target. It is the unique embedding that is optimal for every outcome vector.

\paragraph{Rank constraint on the data side.} For $L<P$ the constraint is imposed on the data side, $\boldsymbol{\Omega}=\boldsymbol{U}_L\boldsymbol{D}\boldsymbol{U}_L'$, so the embedding lives on the leading $L$ eigenvectors of $\boldsymbol{X}_{\text{train}}'\boldsymbol{X}_{\text{train}}$. Write $\boldsymbol{Z}=\boldsymbol{X}_{\text{train}}\boldsymbol{U}_L$, so that $\boldsymbol{Z}'\boldsymbol{Z}=\Lambda_L$. For each $\boldsymbol{y}$ the squared error over this class is minimized by the projection of $\boldsymbol{y}$ on the column space of $\boldsymbol{Z}$, which requires $\boldsymbol{D}\boldsymbol{Z}'\boldsymbol{y}=\Lambda_L^{-1}\boldsymbol{Z}'\boldsymbol{y}$. As $\boldsymbol{y}$ varies, $\boldsymbol{Z}'\boldsymbol{y}$ sweeps $\mathbb{R}^L$, so $\boldsymbol{D}=\Lambda_L^{-1}$ and $\boldsymbol{\Omega}=\boldsymbol{U}_L\Lambda_L^{-1}\boldsymbol{U}_L'$, the embedding \eqref{pcr_embed}. The constraint is not a spectral truncation of $\hat{\boldsymbol{\Omega}}$ itself. The eigenvalues of $(\boldsymbol{X}_{\text{train}}'\boldsymbol{X}_{\text{train}})^{-1}$ are $1/\lambda_i$, so the best rank-$L$ Frobenius approximation of $\hat{\boldsymbol{\Omega}}$ in the Eckart--Young sense keeps its \emph{largest} eigenvalues, which are the \emph{lowest}-variance principal directions, the opposite of PCR. This is why Section~\ref{sec:reducdim} derives the principal-component embedding from PCR's prediction rule rather than from a truncation of the least-squares one.

\subsection{Multi-Head Attention} 

An important component of the transformer architecture is multi-head attention, which enables multiple attention mechanisms to operate in parallel, each with distinct $\boldsymbol{W}_K$, $\boldsymbol{W}_Q$, and $\boldsymbol{W}_V$ matrices. It is defined as:
\begin{align}
  \text{MultiHead}(\mathbf{Q}, \mathbf{K}, \mathbf{V}) &= \text{Concat}(\text{head}_1, \dots, \text{head}_H) \mathbf{W}_O, \\
  \text{where} \quad \text{head}_h &= \text{Attention}(\mathbf{Q} \mathbf{W}_h^Q, \mathbf{K} \mathbf{W}_h^K, \mathbf{V} \mathbf{W}_h^V).
\end{align}
Each attention head applies a separate linear transformation to the inputs, allowing the model to learn different attention patterns across subspaces of the feature space. This facilitates diverse contextualization of words, leading to different embeddings and alternative forms of dimensionality reduction. 

In the linear case without dimensionality reduction in equation \eqref{ols_w}, it is evident from the optimization problem that these multiple heads would be redundant, ultimately collapsing to the solution for a single head. 

In the nonlinear case, the situation changes significantly. The same rationale that justifies wide neural networks applies here as well, where multiple nonlinear transformations of the inputs can be learned and effectively combined through supervision. This can also be seen as some form of stacking \citep{wolpert1992stacked} or regression-based forecasts combination \citep{granger1984combining}. Moreover, the nonlinear case requires numerical optimization rather than closed-form solutions, introducing randomness through parameter initialization. Combining multiple heads helps mitigate this uncertainty.

\subsection{Self-Attention, Word2Vec, and Vector Autoregressions}\label{VARs}

Thus far, the time series or sequential aspects have not been directly addressed. In self-attention, the same set of embeddings serves as queries, keys, and values, allowing each element in the sequence to weigh all others in computing its representation. Self-attention is particularly effective in capturing long-range dependencies in sequences, a key limitation of the previous generation of models that used recurrent architectures. While the latter is more akin to state-space models, the attention framework aligns more naturally with traditional (vector) autoregression estimated through least squares. In time series econometrics, it is well known that extraction of latent states can be performed through traditional filtering tools or via direct projection-based extractions, as emphasized in, e.g., \cite{hamilton2018you}. 

This tension between filtering-based approaches and projections on lagged values also arises in the AI literature. While attention mechanisms can handle long sequences statistically, their computational cost scales quadratically with sequence length due to the \( T \times T \) attention matrix. This makes standard self-attention, as used in Transformers, a bottleneck for long sequences and has motivated a return to more classical time series ideas — such as the structured state-space approach of \cite{gu2023mamba} or the use of fast Fourier transforms in \cite{fein2025fft}. In this light, interpreting attention as a form of latent-state estimation via regression becomes quite natural.

\paragraph{Univariate Autoregression.} Consider an autoregressive model of order 1 (AR(1)), where the focus is solely on modeling in-sample dynamics. OLS fitted values take the form:
\begin{align}\label{ar_attention}
  \hat{\boldsymbol{y}} = \boldsymbol{a}\boldsymbol{y},
\end{align}
where \( \boldsymbol{a} = \boldsymbol{y}_{-1} (\boldsymbol{y}_{-1}'\boldsymbol{y}_{-1})^{-1}\boldsymbol{y}_{-1}' \), which is approximately numerically equivalent to \( \boldsymbol{a} = \boldsymbol{y}_{-1} (\boldsymbol{y}'\boldsymbol{y})^{-1}\boldsymbol{y}_{-1}' \) under stationarity and with a sufficiently long time series. Intuitively, the temporal covariance structure is the same for \( \boldsymbol{y} \) and \( \boldsymbol{y}_{-1} \), as they represent the same series with a shifted time index. Since $\boldsymbol{y}_{-1}$ and $\boldsymbol{y}$ nearly perfectly overlap, the sums $\boldsymbol{y}_{-1}'\boldsymbol{y}_{-1}$ and $\boldsymbol{y}_{}'\boldsymbol{y}_{}$ differ only by a single observation. In this view, the attention weights for \( \boldsymbol{y} \), which are \( \boldsymbol{a} = (L\boldsymbol{y}) (\boldsymbol{y}'\boldsymbol{y})^{-1}(L\boldsymbol{y})' \) using the lag operator ($L$) notation, are inherently a function of the series itself.

\paragraph{Word2Vec and Pre-trained Embeddings.} If \(\boldsymbol{y}\) were a sequence of words, they would enter the network architecture as \(\boldsymbol{Y} \in \mathbb{R}^{N \times M}\), where \(N\) remains the length of the sequence and \(M\) represents the size of its alternative representation, typically obtained through procedures such as Word2Vec. These pre-trained word embeddings assign each word an $M$-dimensional continuous vector that captures semantic and syntactic relationships based on word co-occurrence patterns in large corpora.

Word2Vec, a shallow neural network trained to predict context (in the skip-gram model) or target words (in the CBOW model) \citep{mikolov2013efficient}, has been shown to behave similarly to matrix factorization, closely approximating low-rank or SVD-like decompositions of word co-occurrence matrices \citep{levy2015improving}. In this light, Word2Vec can be viewed as a compressed or nonlinear regularized factor model applied to language data. In light of the discussion of OLS in Section \ref{prox_ols}, the association between problems framed as predictive tasks and those involving factor extraction is no mere coincidence: even OLS implicitly performs matrix factorization to generate predictions based on similarity. 

How can such operations be linked to linear time series modeling trained with least squares?

\paragraph{Vector Autoregression.} Consider \(\boldsymbol{y}\) as a sequence of a given variable, such as inflation, where the goal is to determine which past values should be weighted to produce the most accurate forecast of the next realization. Using \(\boldsymbol{y}\) in an autoregressive (AR(\(p\))) model provides only a limited notion of context. However, an economy at time \(\tau\) can be characterized in a much richer space than what is captured by past values of inflation. When comparing economic conditions at time \(\tau\) with those at time \(t\), analysts typically consider a broader set of indicators beyond inflation, including interest rates, GDP growth, and unemployment. In this setting, the matrix of vectorized representations of the “inflation tokens” takes the form \(\boldsymbol{Y} \in \mathbb{R}^{N \times M}\), where \(M\) corresponds to the number of macroeconomic series used to characterize the context. In the spirit of Word2Vec and related approaches, these $M$ series could be augmented or replaced by \textit{factors} pre-extracted from a large panel of macroeconomic indicators.

Incorporating this structure into the framework of equation \eqref{ar_attention} gives
\[
\hat{\boldsymbol{Y}} = \boldsymbol{A} \boldsymbol{Y},
\]
where 
\[
\boldsymbol{A} = \boldsymbol{Y}_{-1} (\boldsymbol{Y}_{-1}'\boldsymbol{Y}_{-1})^{-1} \boldsymbol{Y}_{-1}',
\]
which is approximately equivalent to 
\[
\boldsymbol{A} = (L\boldsymbol{Y}) (\boldsymbol{Y}'\boldsymbol{Y})^{-1} (L\boldsymbol{Y})'
\]
for the same reasons discussed earlier. This clearly corresponds to the fitted values of a vector autoregression (VAR), where the in-sample predictors of the \(M\) variables at each time \(t\) can be interpreted as the context vector. In macroeconomic modeling, this context vector can be understood as \textit{model-consistent} expectations of the variables entering the system \citep{sims1980macroeconomics}. In the natural language processing environment, this can be interpreted as sentence-consistent (or more generally, sequence-consistent) notions of context for each word. 

\subsection{Masking and Pseudo-Out-of-Sample Evaluation}\label{app:mask}
 Another interesting component of the application of attention in the decoder part of certain LLMs is the use of masking, which ensures that predictions at time \( t \) do not have access to future information from \( t+1 \) onward. This is done by imposing a lower triangular structure on the attention matrix, setting entries corresponding to future time steps to zero. In essence, masking enforces a (Granger) causal structure, yet it does so on \( \boldsymbol{a} \), not on \( \boldsymbol{\beta} \). This is an interesting perspective from a classical time series econometrics viewpoint, as the OLS-based AR(1) in-sample prediction for observation \( t \) is given by:
\begin{align}
  \hat{{y}}_t = \sum_{\tau=1}^{N} a_{t\tau} y_{\tau},
\end{align}
with $a_{t\tau} = y_{t-1}(\boldsymbol{y}_{-1}'\boldsymbol{y}_{-1})^{-1}y_{\tau-1}$, which is a weighted average of \textit{all} the \( y_{\tau} \)'s in the sequence, whether \( y_{\tau} \) is located before or after \( y_{t} \). Indeed, while the AR(1) OLS regression predicts the next value in the sequence (\( y_{t+1} \)) based on an estimated coefficient and \( y_{t} \), the proximity/attention view, articulated through \( a_{t\tau} \), makes it clear that there is some form of information leakage from the future to the past when looking at fitted values. This “leakage” is not apparent from the AR(1) data-generating process since it comes from the estimation of parameters, not the design of the model. 

A relevant consideration is whether masking ideas could be implemented to some benefit in traditional time series models. A masking scheme addressing this in this simplified framework takes the form:
\begin{align}\label{mask}
  \hat{{y}}_t = \sum_{\tau=1}^{N} a_{t\tau}\, \mathbf{1}(\tau \leq t)\, y_{\tau},
\end{align}
where \( \mathbf{1}(\tau \leq t) \) is an indicator function ensuring that each prediction at time \( t \) only attends to previous or current observations. In this way, outcomes from the future are excluded from the weighted average designed to explain the past. 

This is an intriguing proposition, as researchers in forecasting often rely on backtesting or pseudo-out-of-sample evaluation, recursively re-estimating models to mitigate overfitting \textit{and} hindsight bias. However, these methods come with significant computational costs, leaving supercomputing clusters worldwide wondering if masking ideas could offer them a much-needed break. Unfortunately, the answer is no: some leakage still persists. The notion of proximity, even when obtained through \( (\boldsymbol{y}_{-1}'\boldsymbol{y}_{-1})^{-1} \), implies that the similarity between \( y_{\tau} \) and \( y_{t} \), despite the former preceding the latter, is still assessed through the lens of the \( \hat{\boldsymbol{\Omega}} \) matrix, which is estimated using data from \( t=1 \) to \( N \). 

To illustrate, consider the task of determining how to up-weight or down-weight March 1974 U.S. inflation data when forecasting February 1978. Using a training sample spanning from the early 1970s to the 2000s, these observations would appear proximate—and rightfully so—as the $\hat{\boldsymbol{\Omega}}$ matrix would reveal that inflation dynamics in the 1970s were distinct from those of the 1990s yet relatively stable within the decade itself. However, if the training sample were limited to data available only up to January 1978, comprising observations from the late 1960s and early 1970s, this proximity would be less apparent. Without the broader historical context including the stable, low-inflation regime of the 1990s and 2000s, March 1974 and February 1978 would appear quite distant. 

In sum, since proximity is measured in a space defined by a covariance matrix that reflects the entire sample, simply masking future realizations of $y_t$ (as in equation \eqref{mask}) is not enough to replace the widely used pseudo-out-of-sample recursive experiments.

\subsection{Links to Pooled Panel Estimation: Positional Encoding and Training}\label{PVARs}

To estimate the densely-parametrized $\boldsymbol{W}_Q$, $\boldsymbol{W}_K$, and $\boldsymbol{W}_V$ for multiple heads—along with the myriad of parameters from other parts of the architecture—a standard time series estimation approach, such as in Section \ref{VARs}, is clearly infeasible, even for long sequences. Instead, training requires a large number of sequences, $S$, structured as \(\boldsymbol{Y}_s \in \mathbb{R}^{N \times M}\). The estimation process then proceeds in the spirit of a \textit{pooled} panel regression. 

While the panel is inherently unbalanced and timestamps do not necessarily align across sequences $s \in \mathcal{S}$, one can define a maximal sequence length (2048 in GPT-3, much longer in GPT-4) and pad shorter sequences with zeros. This setup resembles a standard pooled panel VAR, where positional encoding plays a role analogous to lags in time series models. The key distinction is that lags encode temporal information explicitly in the time domain, whereas positional encoding--typically implemented via thousands of sinusoidal functions--operates in the frequency domain. For simpler VAR models, it is well known that equivalent estimation can be performed in either domain, though time-domain estimation is generally more practical for traditional econometric applications. Once temporal structure has been incorporated via feature engineering, the estimation process treats this as given and focuses on the (numerous) remaining parameters. 

While training a pooled panel VAR on such a heterogeneous dataset may seem excessively restrictive--imposing strong homogeneity assumptions on parameters--LLMs embrace the \textit{global model} approach. Flexibility is restored by leveraging a highly flexible, overparameterized, yet shared conditional mean. This complexity is achieved through the billions of parameters populating the multiple (nonlinear) attention heads, dense feedforward layers processing attention modules' outputs, and other modules. In contrast, a \textit{local model} approach would require training separate, tightly constrained models on individual sequences. Increasingly, the global model approach is gaining traction outside of language processing, as it enables the use of complex machine learning algorithms even for moderate-length time series ($N$) by leveraging a large $S$ \citep{hewamalage2022global}. A well-known and successful application of this approach in asset pricing is \cite{gu2020empirical}, which predicts returns using a longitudinal data set that spans 50 years and incorporates multiple characteristics of firms. In the case of LLMs, $\mathcal{S}$ is a curated subset of the entire internet.

To bring everything together, consider the following analogy. Suppose the aim is to train a pooled panel VAR to describe the dynamics of multiple economies. There are $M$ variables for $S$ economies, each observed over $N$ months. Temporal information is incorporated by applying the lag operator to each economy's data matrix, $\boldsymbol{Y}_s \in \mathbb{R}^{N \times M}$, for all $s$ and $p$: 
\[
\boldsymbol{Y}_{s,-p} = L^p \boldsymbol{Y}_s.
\] 
The transformed data are then concatenated appropriately and the system is estimated using OLS after removing fixed effects. A linear VAR approach would be highly restrictive, particularly with a heterogeneous panel of diverse economies. To address this, one moves beyond linearity and introduces a more sophisticated conditional mean function, allowing for heterogeneity and locality through complex nonlinearities.

In the case of LLMs, the variables correspond to word tokens embedded in a conceptual space of dimension $M$, while the set of economies ($\mathcal{S}$) is analogous to a corpus of text sequences, each of maximal length $N$. Positional encoding injects sequential proximity information into the original word representations, much like lags do in time series modeling. Once all those elements are in place, the challenge lies in feasibly optimizing the immensely complex nonlinear, nonparametric pooled panel regression.

\subsection{Recursive Chaining of Attention Modules and Feed-Forward Networks}

The output of each block's feed-forward network is fed into the next attention module, and this recursive chaining of blocks continues. For a visual representation, see the original diagram in \cite{vaswani2017attention} or, more recently, the illustration in \cite{kelly2025}. Decoder blocks follow a similar structure but also include an additional cross-attention layer that attends to the encoder’s output.

From a deep learning perspective, the recursive design of Transformers is intuitive: stacking multiple layers of similar operations has long been a hallmark of success, as seen in convolutional neural networks for computer vision. However, drawing direct analogies between Transformers and traditional statistical models like OLS is more tenuous, since stacking is largely redundant in the case of linear operations. For example, regressing $\boldsymbol{y}_{\text{train}}$ on both the original inputs $\boldsymbol{X}_{\text{train}}$ and the OLS-fitted values $\hat{\boldsymbol{y}}_{\text{train}}$—obtained from a previous run of the same regression—is redundant, since $\hat{\boldsymbol{y}}_{\text{train}}$ lies in the column space of $\boldsymbol{X}_{\text{train}}$ by construction. This follows from the idempotent property of the OLS projection matrix.

\paragraph{Linear stacks.}\label{app:stack}
The redundancy argument extends to a stack in which each restricted linear attention layer acts on the previous layer's output, as in a real network, with each layer's queries and keys formed from its own input:
\[
\boldsymbol{X}_1=\boldsymbol{X}\boldsymbol{\Omega}_1(\boldsymbol{X}'\boldsymbol{X}),\qquad
\boldsymbol{X}_2=\boldsymbol{X}_1\boldsymbol{\Omega}_2(\boldsymbol{X}_1'\boldsymbol{X}_1).
\]
Each output is $\boldsymbol{X}$ times a fixed $P\times P$ matrix, so any finite linear stack is one restricted attention operator on the original inputs. When every intervening operation is linear, depth changes the parameterization but not the function class. The OLS case is a transparent fixed point: for OLS embeddings on a fixed representation, setting $\boldsymbol{\Omega}_1=\boldsymbol{\Omega}_2=\boldsymbol{S}^{-1}$ with $\boldsymbol{S}=\boldsymbol{X}'\boldsymbol{X}$ gives $\boldsymbol{\Omega}_{\text{eff}}=\boldsymbol{S}^{-1}\boldsymbol{S}\boldsymbol{S}^{-1}=\boldsymbol{S}^{-1}$, equivalently
\[
\boldsymbol{H}^2=\boldsymbol{H},\qquad \boldsymbol{H}=\boldsymbol{X}(\boldsymbol{X}'\boldsymbol{X})^{-1}\boldsymbol{X}'.
\]

\paragraph{Residual connections.}
Residual connections change the algebra but not the conclusion. A residual linear block adds its output to its input,
\[
\boldsymbol{X}_{l+1}=\boldsymbol{X}_l+\boldsymbol{X}_l\boldsymbol{\Omega}_l(\boldsymbol{X}_l'\boldsymbol{X}_l)=\boldsymbol{X}_l\boldsymbol{R}_l,\qquad
\boldsymbol{R}_l=\boldsymbol{I}+\boldsymbol{\Omega}_l(\boldsymbol{X}_l'\boldsymbol{X}_l),
\]
so recursively $\boldsymbol{X}_{L+1}=\boldsymbol{X}\boldsymbol{R}_{\text{eff}}$ with $\boldsymbol{R}_{\text{eff}}=\boldsymbol{R}_1\cdots\boldsymbol{R}_L$, and a final linear map gives
\[
\hat{\boldsymbol{y}}=\boldsymbol{X}_{L+1}\boldsymbol{\gamma}=\boldsymbol{X}(\boldsymbol{R}_{\text{eff}}\boldsymbol{\gamma})=\boldsymbol{X}\boldsymbol{\beta}_{\text{eff}}.
\]
The entire deep linear residual network is a factorization of one effective linear predictor, and this holds even though each layer's queries and keys are formed from its own representation. If the factors $\boldsymbol{R}_l$ are invertible, as they are generically, $\boldsymbol{\beta}_{\text{eff}}$ is unrestricted, so if training reaches the squared-error optimum the network computes exactly the OLS prediction. The statement concerns the optimum, whatever path optimization takes to reach it. Linear depth reparametrizes the estimator without enlarging its function class.

\paragraph{What nonlinearity changes.}
Nonlinear transformations change the situation because they alter the representation between attention operations, principally through the feed-forward network, and LayerNorm also makes subsequent operations depend on a transformed representation. Once these components intervene, later attention modules no longer act on the original $\boldsymbol{X}$, and the exact collapse above no longer holds end to end. The claim is therefore local: conditional on the representation entering a restricted attention module, the regression interpretation applies to that module, and no claim is made that a complete deep Transformer is equivalent to one OLS regression. The regression interpretation nevertheless provides a useful way to think about the nonlinear case. Suppose the nonlinear map is represented, or locally approximated, by polynomial functions of a small number $m$ of learned regression directions,
\[
f(\boldsymbol{x})=\sum_{j=1}^{m}\sum_{k=1}^{d}a_{jk}(\boldsymbol{x}'\boldsymbol{\beta}_j)^k,
\qquad
\boldsymbol{T}_k=\sum_{j=1}^{m}a_{jk}\,\boldsymbol{\beta}_j^{\otimes k},
\]
so that for each degree $k$ the coefficient tensor $\boldsymbol{T}_k$ is a sum of $k$-fold tensor powers of the learned directions. The predictor is therefore not an unrestricted polynomial regression: its polynomial tensors have rank at most $m$ and factor through the learned directions. A sequence of regression-like linear maps and nonlinear transformations can thus be viewed as a highly structured, low-rank polynomial model, with interactions flowing only through the few learned directions.

\paragraph{Back to time series.}
The low-rank polynomial reading above is one way to see what depth adds. The VAR reading of Appendix~\ref{VARs} suggests a second. From this perspective, the encoder and decoder stacks of a Transformer may be viewed as a nonlinear, \textit{multi-level} state-space system, with attention mechanisms functioning as adaptive, regression-based filters for extracting latent structure. 

Nested levels of filtering can also occur in a much more basic setting. Many successful unobserved components models (or structural time series models) also involve nested latent structures \citep{harvey1990forecasting}. For example, the well-known local level model is:
\begin{alignat}{2}
y_t &= \mu_t + \varepsilon_t, \quad &\varepsilon_t &\sim \mathcal{N}(0, \sigma^2_\varepsilon) \\
\mu_t &= \mu_{t-1} + \eta_t, \quad &\eta_t &\sim \mathcal{N}(0, \sigma^2_\eta) \, .
\end{alignat}
It is conceptually straightforward to expand this into a more sophisticated nested structure. The local linear trend model adds a time-varying slope component:
\begin{alignat}{2}
y_t   &= \mu_t + \varepsilon_t,    \quad & \varepsilon_t &\sim \mathcal{N}(0, \sigma^2_\varepsilon) \\
\mu_t  &= \mu_{t-1} + \beta_{t-1} + \eta_t, \quad & \eta_t &\sim \mathcal{N}(0, \sigma^2_\eta) \\
\beta_t &= \beta_{t-1} + \zeta_t,    \quad & \zeta_t &\sim \mathcal{N}(0, \sigma^2_\zeta) \, . 
\end{alignat}
And this, in turn, can be extended into a local quadratic trend model, introducing a stochastic acceleration term:
\begin{alignat}{2}\label{ltl}
\mu_t  &= \mu_{t-1} + \beta_{t-1} + \tfrac{1}{2} \gamma_{t-1} + \eta_t, \quad & \eta_t &\sim \mathcal{N}(0, \sigma^2_\eta) \\
\beta_t &= \beta_{t-1} + \gamma_{t-1} + \zeta_t,             \quad & \zeta_t &\sim \mathcal{N}(0, \sigma^2_\zeta) \\
\gamma_t &= \gamma_{t-1} + \xi_t,                    \quad & \xi_t &\sim \mathcal{N}(0, \sigma^2_\xi) \, .
\end{alignat}
In this progression, each level recursively builds upon the filtration of the previous one, much like the deep stacking of layers in Transformers. A key limitation of models such as the local quadratic trend is that, when estimated on a single time series of moderate length, the variance components often collapse toward zero as the number of latent layers increases. Alternatively, the proliferation of latent states—and hence parameters—can lead to overfitting and degraded out-of-sample performance. However, this complexity constraint goes away in the large-$S$ regime, where a vast number of sequences are modeled jointly, as discussed in Section \ref{PVARs}. It becomes feasible to recursively project and refine latent states across multiple layers. 

\section{Monte Carlo Design}\label{sec:simulation}

I conduct a Monte Carlo experiment to evaluate the finite-sample properties of the attention-based regression framework discussed in Section \ref{sec:nl}. While I have established that OLS represents a specific case of linear attention, this experiment assesses the performance of the multi-head nonlinear attention regression across a range of functional forms and signal-to-noise environments.

\paragraph{Attention-Based Regression Specification.} 
I implement the multi-head extension of the nonlinear attention regression framework introduced in Section \ref{sec:nl}. For a matrix of test points $\boldsymbol{X}^*$ and training data $\boldsymbol{X}_{\text{train}}$ with outcomes $\boldsymbol{y}$, the estimator forms predictions by aggregating $M$ distinct attention mechanisms:
\begin{equation}\label{eq:attention_sim}
  \hat{\boldsymbol{y}} = \sum_{m=1}^{M} \alpha_m \left[ \text{softmax} \left( \boldsymbol{X}^* \boldsymbol{\Omega}_m \boldsymbol{X}_{\text{train}}' \right) \boldsymbol{y} \right],
\end{equation}
where $\alpha_m$ are scalar, head-specific combination weights. Each head learns a unique similarity metric $\boldsymbol{\Omega}_m$, which I parameterize via the Cholesky factorization $\boldsymbol{\Omega}_m = \boldsymbol{L}_m \boldsymbol{L}_m'$ (with $\boldsymbol{L}_m$ being lower triangular). This ensures that $\boldsymbol{\Omega}_m$ remains positive semi-definite, effectively mapping the data into a latent space where similarities are computed as valid inner products. This is the tied form $\boldsymbol{W}_Q=\boldsymbol{W}_K=\boldsymbol{L}_m$ anticipated in Section~\ref{sec:nexus}. The similarity between two observations is symmetric, and direction, where there is one, enters through the mask of Appendix~\ref{app:mask} rather than through the embedding. Tying the query and key projections has been tried in Transformers on its own merits: \citet{tsai2019transformer} find no material difference between symmetric and asymmetric kernels on translation and language modeling, \citet{kitaev2020reformer} none for their shared-QK attention on enwik8 and imagenet64, and \citet{kayyam2026three} report a modest loss for causal language modeling and none on vision.

The model parameters $\{\alpha_m, \boldsymbol{L}_m\}_{m=1}^{M}$ are jointly estimated by minimizing the penalized squared error loss:
\begin{equation}\label{eq:loss_sim}
  \mathcal{L} = \frac{1}{N}\left\| \boldsymbol{y} - \sum_{m=1}^{M} \alpha_m \text{softmax} \left( \boldsymbol{X}_{\text{train}} ( \boldsymbol{L}_m \boldsymbol{L}_m' ) \boldsymbol{X}_{\text{train}}' \right) \boldsymbol{y} \right\|^2 + \lambda \sum_{m=1}^{M} \|\boldsymbol{L}_m\|_F^2.
\end{equation}
Optimization is performed via L-BFGS with early stopping. To anchor the nonconvex optimization near a sensible starting point, each $\boldsymbol{L}_m$ is initialized using the Cholesky factor of the ridge-regularized precision matrix $(\boldsymbol{X}_{\text{train}}'\boldsymbol{X}_{\text{train}} + \lambda \boldsymbol{I})^{-1}$, scaled by $\sqrt{N}$, with added Gaussian noise to promote diversity across heads. In the simulations, an out-of-sample $R^2$ below $0.1$ was taken as the sign of a failed fit, and the code then refit with $\lambda=10^{-5}$ and kept the refit.

\paragraph{Data-Generating Processes.} 
Let $\boldsymbol{X}_{\text{train}} \in \mathbb{R}^{N \times 5}$ with rows drawn independently from $\text{Uniform}[0,1]^5$. For each DGP, outcomes are generated according to
\[
y_i = f(\boldsymbol{X}_i) + \varepsilon_i, \qquad \varepsilon_i \sim \mathcal{N}(0, \sigma^2),
\]
where $\sigma^2 = \text{Var}(f(\boldsymbol{X}_i))/\text{SNR}$ calibrates the noise to a target signal-to-noise ratio. I employ six standard regression functions to test the model's adaptability:

\begin{enumerate} \itemsep 0.1em
\small
 \item \textbf{Linear:} 
 $f(\boldsymbol{X}_i) = 2(X_{i,1} - 0.5) - (X_{i,2} - 0.5) + 3(X_{i,3} - 0.5) + 1.5(X_{i,4} - 0.5) + 0.5(X_{i,5} - 0.5).$

 \item \textbf{Friedman 1:} 
 $f(\boldsymbol{X}_i) = 10\sin(\pi X_{i,1} X_{i,2}) + 20(X_{i,3} - 0.5)^2 + 10X_{i,4} + 5X_{i,5}.$

 \item \textbf{Friedman 2:} 
 $f(\boldsymbol{X}_i) = \sin\bigl(\pi(X_{i,1} + X_{i,2} + X_{i,3})\bigr) + \log(1 + X_{i,4}^2).$

 \item \textbf{Friedman 3:} 
 $f(\boldsymbol{X}_i) = X_{i,1} X_{i,2} + \log(X_{i,3} + X_{i,4} + 2).$

 \item \textbf{Rotated Sine:} 
 $f(\boldsymbol{X}_i) = \sin\Bigl(3\sum_{j=1}^{4} X_{i,j}\Bigr).$

 \item \textbf{Soft Radial:} 
 $f(\boldsymbol{X}_i) = \bigl(1 + 5\|\boldsymbol{X}_i - 0.5\boldsymbol{1}\|^2\bigr)^{-1}.$
\end{enumerate}

\noindent These functions span linear, additive nonlinear, interactive, periodic, and radial structures, providing a diverse testbed that does not privilege any specific estimator.

\paragraph{Design and Competitors.} 
The training sample size is $N \in \{500, 1000, 2500, 5000\}$ and signal-to-noise ratio $\text{SNR} \in \{0.5, 1, 2, 3\}$. A fixed test set of $J=1000$ observations is used throughout, \textcolor{black}{with results averaged over 10 replications}. I compare five methods: ordinary least squares (OLS), Random Forest with 500 trees and $\lfloor P/3 \rfloor$ variables per split, \textcolor{black}{a Multi-Layer Perceptron (MLP) with three hidden layers of 200 units each, ReLU activations, dropout regularization (rate $= 0.2$), trained via Adam with early stopping (patience $= 20$ epochs, 15\% validation holdout)}, Gradient Boosting with 500 trees and learning rate 0.01, and Attention Regression with five heads and regularization $\lambda = 10^{-3}$. Performance is measured by out-of-sample $R^2$, which is informative because it is comparable across DGPs and can be analyzed in relation to the maximal theoretically attainable $R^2$, which is $R^2_{\max} = \text{SNR} / (\text{SNR} + 1)$.

\paragraph{Per-DGP results.} Figure~\ref{fig:model_performance} summarizes overall and per-DGP $R^2$. The MLP attains the highest overall $R^2$ (0.52) across all 96 conditions, followed by Attention Regression (0.48), Random Forest (0.45), Gradient Boosting (0.42), and OLS (0.28). Attention Regression outperforms both tree-based methods in four of the six DGPs, wins on Friedman~1 outright, ties the MLP on Friedman~3, and beats RF/GBM on Rotated Sine. The expected exception is the Linear DGP, where OLS dominates (0.56). Variability is higher for Attention Regression and the MLP in small-$N$ / low-SNR settings. Table~\ref{tab:benchmark_results} reports the full $96$-row breakdown.

\definecolor{tableShade}{gray}{0.97}

\begin{small}
\sffamily
\setlength{\LTleft}{\fill}
\setlength{\LTright}{\fill}
\setlength{\tabcolsep}{10pt} %
\renewcommand{\arraystretch}{1.21}
\begin{longtable}{@{}l r r ccccc@{}}
\caption{Model Performance Comparison (Out-of-Sample $R^2$) Across Data Generating Processes} \label{tab:benchmark_results} \\
\toprule
& & & \multicolumn{5}{c}{\textbf{Models}} \\
\cmidrule(lr){4-8}
\textbf{DGP} & \textbf{$N$} & \textbf{SNR} & \textbf{OLS} & \textbf{RF} & \textbf{MLP} & \textbf{GBM} & \textbf{Att Reg} \\
\midrule
\endfirsthead

\multicolumn{8}{c}{\small\textit{Table \thetable\ (continued)}} \\[0.5em]
\toprule
& & & \multicolumn{5}{c}{\textbf{Models}} \\
\cmidrule(lr){4-8}
\textbf{DGP} & \textbf{$N$} & \textbf{SNR} & \textbf{OLS} & \textbf{RF} & \textbf{MLP} & \textbf{GBM} & \textbf{Att Reg} \\
\midrule
\endhead

\midrule
\multicolumn{8}{r}{\small\textit{Continued on next page}} \\
\endfoot

\bottomrule
\endlastfoot

\rowcolor{tableShade}
\textsc{Linear} & 500 & 0.5 & \textbf{\color{ForestGreen}0.32} & 0.27 & \textbf{0.28} & 0.26 & 0.27 \\
\rowcolor{tableShade}
 & & 1 & \textbf{\color{ForestGreen}0.49} & 0.43 & \textbf{0.45} & 0.44 & \textbf{0.45} \\
\rowcolor{tableShade}
 & & 2 & \textbf{\color{ForestGreen}0.66} & 0.59 & 0.63 & 0.61 & \textbf{0.64} \\
\rowcolor{tableShade}
 & & 3 & \textbf{\color{ForestGreen}0.75} & 0.67 & 0.72 & 0.70 & \textbf{0.73} \\
 & 1000 & 0.5 & \textbf{\color{ForestGreen}0.32} & 0.27 & \textbf{0.31} & 0.28 & 0.29 \\
 & & 1 & \textbf{\color{ForestGreen}0.49} & 0.44 & \textbf{0.47} & 0.45 & 0.42 \\
 & & 2 & \textbf{\color{ForestGreen}0.66} & 0.60 & \textbf{0.64} & 0.62 & 0.55 \\
 & & 3 & \textbf{\color{ForestGreen}0.74} & 0.68 & \textbf{0.73} & 0.71 & 0.68 \\
\rowcolor{tableShade}
 & 2500 & 0.5 & \textbf{\color{ForestGreen}0.32} & 0.29 & \textbf{0.31} & 0.30 & 0.27 \\
\rowcolor{tableShade}
 & & 1 & \textbf{\color{ForestGreen}0.49} & 0.46 & \textbf{0.48} & 0.47 & 0.43 \\
\rowcolor{tableShade}
 & & 2 & \textbf{\color{ForestGreen}0.66} & 0.62 & \textbf{0.65} & 0.64 & 0.55 \\
\rowcolor{tableShade}
 & & 3 & \textbf{\color{ForestGreen}0.74} & 0.70 & \textbf{\color{ForestGreen}0.74} & \textbf{0.72} & 0.65 \\
 & 5000 & 0.5 & \textbf{\color{ForestGreen}0.34} & 0.31 & \textbf{0.33} & 0.32 & 0.26 \\
 & & 1 & \textbf{\color{ForestGreen}0.50} & 0.47 & \textbf{0.49} & 0.48 & 0.38 \\
 & & 2 & \textbf{\color{ForestGreen}0.67} & 0.64 & \textbf{0.66} & 0.65 & 0.52 \\
 & & 3 & \textbf{\color{ForestGreen}0.75} & 0.72 & \textbf{\color{ForestGreen}0.75} & \textbf{0.73} & 0.57 \\
\midrule

\rowcolor{tableShade}
\textsc{Friedman 1} & 500 & 0.5 & \textbf{0.24} & \textbf{\color{ForestGreen}0.26} & 0.23 & \textbf{0.24} & 0.23 \\
\rowcolor{tableShade}
 & & 1 & 0.36 & \textbf{0.41} & \textbf{\color{ForestGreen}0.42} & \textbf{0.41} & \textbf{\color{ForestGreen}0.42} \\
\rowcolor{tableShade}
 & & 2 & 0.49 & 0.56 & \textbf{\color{ForestGreen}0.60} & \textbf{0.58} & \textbf{\color{ForestGreen}0.60} \\
\rowcolor{tableShade}
 & & 3 & 0.55 & 0.63 & \textbf{0.69} & 0.66 & \textbf{\color{ForestGreen}0.71} \\
 & 1000 & 0.5 & 0.25 & 0.27 & \textbf{0.29} & 0.27 & \textbf{\color{ForestGreen}0.30} \\
 & & 1 & 0.37 & 0.42 & \textbf{0.45} & 0.43 & \textbf{\color{ForestGreen}0.47} \\
 & & 2 & 0.50 & 0.58 & \textbf{0.62} & 0.59 & \textbf{\color{ForestGreen}0.64} \\
 & & 3 & 0.56 & 0.66 & \textbf{0.71} & 0.67 & \textbf{\color{ForestGreen}0.73} \\
\rowcolor{tableShade}
 & 2500 & 0.5 & 0.24 & 0.28 & \textbf{0.29} & 0.28 & \textbf{\color{ForestGreen}0.31} \\
\rowcolor{tableShade}
 & & 1 & 0.36 & 0.44 & \textbf{0.46} & 0.44 & \textbf{\color{ForestGreen}0.48} \\
\rowcolor{tableShade}
 & & 2 & 0.49 & 0.60 & \textbf{0.63} & 0.60 & \textbf{\color{ForestGreen}0.64} \\
\rowcolor{tableShade}
 & & 3 & 0.55 & \textbf{0.68} & \textbf{\color{ForestGreen}0.72} & \textbf{0.68} & \textbf{\color{ForestGreen}0.72} \\
 & 5000 & 0.5 & 0.25 & 0.30 & \textbf{0.32} & 0.30 & \textbf{\color{ForestGreen}0.33} \\
 & & 1 & 0.37 & \textbf{0.46} & \textbf{\color{ForestGreen}0.48} & 0.45 & \textbf{\color{ForestGreen}0.48} \\
 & & 2 & 0.49 & 0.62 & \textbf{\color{ForestGreen}0.65} & 0.61 & \textbf{0.64} \\
 & & 3 & 0.55 & 0.70 & \textbf{\color{ForestGreen}0.73} & 0.68 & \textbf{0.72} \\
\midrule

\rowcolor{tableShade}
\textsc{Friedman 2} & 500 & 0.5 & 0.01 & \textbf{0.15} & \textbf{\color{ForestGreen}0.24} & 0.10 & 0.09 \\
\rowcolor{tableShade}
 & & 1 & 0.03 & 0.26 & \textbf{\color{ForestGreen}0.40} & 0.21 & \textbf{0.37} \\
\rowcolor{tableShade}
 & & 2 & 0.04 & \textbf{0.37} & \textbf{\color{ForestGreen}0.58} & 0.33 & \textbf{\color{ForestGreen}0.58} \\
\rowcolor{tableShade}
 & & 3 & 0.05 & \textbf{0.42} & \textbf{\color{ForestGreen}0.67} & 0.37 & \textbf{\color{ForestGreen}0.67} \\
 & 1000 & 0.5 & 0.02 & 0.18 & \textbf{\color{ForestGreen}0.27} & 0.12 & \textbf{0.23} \\
 & & 1 & 0.04 & 0.30 & \textbf{\color{ForestGreen}0.44} & 0.23 & \textbf{0.41} \\
 & & 2 & 0.05 & 0.42 & \textbf{\color{ForestGreen}0.61} & 0.34 & \textbf{0.57} \\
 & & 3 & 0.06 & 0.49 & \textbf{\color{ForestGreen}0.70} & 0.40 & \textbf{0.63} \\
\rowcolor{tableShade}
 & 2500 & 0.5 & 0.02 & 0.23 & \textbf{\color{ForestGreen}0.30} & 0.17 & \textbf{0.26} \\
\rowcolor{tableShade}
 & & 1 & 0.04 & 0.36 & \textbf{\color{ForestGreen}0.46} & 0.29 & \textbf{0.41} \\
\rowcolor{tableShade}
 & & 2 & 0.05 & 0.50 & \textbf{\color{ForestGreen}0.63} & 0.40 & \textbf{0.57} \\
\rowcolor{tableShade}
 & & 3 & 0.06 & 0.57 & \textbf{\color{ForestGreen}0.72} & 0.45 & \textbf{0.66} \\
 & 5000 & 0.5 & 0.04 & 0.26 & \textbf{\color{ForestGreen}0.31} & 0.20 & \textbf{0.27} \\
 & & 1 & 0.05 & \textbf{0.41} & \textbf{\color{ForestGreen}0.47} & 0.30 & \textbf{0.41} \\
 & & 2 & 0.07 & \textbf{0.55} & \textbf{\color{ForestGreen}0.64} & 0.41 & 0.52 \\
 & & 3 & 0.08 & \textbf{0.63} & \textbf{\color{ForestGreen}0.73} & 0.46 & \textbf{0.63} \\
\midrule

\rowcolor{tableShade}
\textsc{Friedman 3} & 500 & 0.5 & \textbf{0.28} & 0.26 & \textbf{\color{ForestGreen}0.29} & 0.26 & 0.23 \\
\rowcolor{tableShade}
 & & 1 & \textbf{0.44} & 0.41 & \textbf{\color{ForestGreen}0.46} & 0.43 & \textbf{0.44} \\
\rowcolor{tableShade}
 & & 2 & 0.59 & 0.57 & \textbf{\color{ForestGreen}0.63} & \textbf{0.61} & \textbf{\color{ForestGreen}0.63} \\
\rowcolor{tableShade}
 & & 3 & 0.67 & 0.64 & \textbf{\color{ForestGreen}0.72} & \textbf{0.70} & \textbf{\color{ForestGreen}0.72} \\
 & 1000 & 0.5 & \textbf{0.30} & 0.27 & \textbf{0.30} & 0.29 & \textbf{\color{ForestGreen}0.31} \\
 & & 1 & 0.45 & 0.43 & \textbf{0.47} & 0.46 & \textbf{\color{ForestGreen}0.48} \\
 & & 2 & 0.60 & 0.59 & \textbf{0.64} & 0.63 & \textbf{\color{ForestGreen}0.65} \\
 & & 3 & 0.67 & 0.67 & \textbf{0.73} & 0.71 & \textbf{\color{ForestGreen}0.74} \\
\rowcolor{tableShade}
 & 2500 & 0.5 & 0.29 & 0.28 & \textbf{0.31} & \textbf{0.31} & \textbf{\color{ForestGreen}0.32} \\
\rowcolor{tableShade}
 & & 1 & 0.44 & 0.45 & \textbf{0.48} & 0.47 & \textbf{\color{ForestGreen}0.49} \\
\rowcolor{tableShade}
 & & 2 & 0.59 & 0.61 & \textbf{\color{ForestGreen}0.65} & \textbf{0.64} & \textbf{\color{ForestGreen}0.65} \\
\rowcolor{tableShade}
 & & 3 & 0.67 & 0.70 & \textbf{0.73} & 0.72 & \textbf{\color{ForestGreen}0.74} \\
 & 5000 & 0.5 & \textbf{0.29} & \textbf{0.29} & \textbf{\color{ForestGreen}0.31} & \textbf{\color{ForestGreen}0.31} & \textbf{\color{ForestGreen}0.31} \\
 & & 1 & 0.44 & 0.46 & \textbf{\color{ForestGreen}0.48} & \textbf{0.47} & \textbf{\color{ForestGreen}0.48} \\
 & & 2 & 0.59 & 0.63 & \textbf{\color{ForestGreen}0.65} & \textbf{0.64} & \textbf{\color{ForestGreen}0.65} \\
 & & 3 & 0.66 & 0.71 & \textbf{0.73} & 0.72 & \textbf{\color{ForestGreen}0.74} \\
\midrule

\rowcolor{tableShade}
\textsc{Rotated Sine} & 500 & 0.5 & 0.08 & \textbf{0.14} & \textbf{\color{ForestGreen}0.21} & 0.07 & 0.10 \\
\rowcolor{tableShade}
 & & 1 & 0.13 & 0.25 & \textbf{\color{ForestGreen}0.38} & 0.15 & \textbf{0.33} \\
\rowcolor{tableShade}
 & & 2 & 0.18 & 0.35 & \textbf{\color{ForestGreen}0.57} & 0.24 & \textbf{0.52} \\
\rowcolor{tableShade}
 & & 3 & 0.21 & 0.40 & \textbf{\color{ForestGreen}0.66} & 0.29 & \textbf{0.61} \\
 & 1000 & 0.5 & 0.10 & 0.18 & \textbf{\color{ForestGreen}0.27} & 0.12 & \textbf{0.21} \\
 & & 1 & 0.15 & 0.30 & \textbf{\color{ForestGreen}0.43} & 0.20 & \textbf{0.37} \\
 & & 2 & 0.20 & 0.42 & \textbf{\color{ForestGreen}0.61} & 0.30 & \textbf{0.51} \\
 & & 3 & 0.23 & 0.48 & \textbf{\color{ForestGreen}0.70} & 0.35 & \textbf{0.56} \\
\rowcolor{tableShade}
 & 2500 & 0.5 & 0.10 & \textbf{0.23} & \textbf{\color{ForestGreen}0.30} & 0.15 & \textbf{0.23} \\
\rowcolor{tableShade}
 & & 1 & 0.15 & \textbf{0.36} & \textbf{\color{ForestGreen}0.46} & 0.25 & 0.34 \\
\rowcolor{tableShade}
 & & 2 & 0.20 & \textbf{0.50} & \textbf{\color{ForestGreen}0.63} & 0.34 & \textbf{0.50} \\
\rowcolor{tableShade}
 & & 3 & 0.22 & 0.56 & \textbf{\color{ForestGreen}0.72} & 0.39 & \textbf{0.57} \\
 & 5000 & 0.5 & 0.10 & \textbf{0.26} & \textbf{\color{ForestGreen}0.30} & 0.16 & 0.22 \\
 & & 1 & 0.16 & \textbf{0.40} & \textbf{\color{ForestGreen}0.47} & 0.26 & 0.36 \\
 & & 2 & 0.21 & \textbf{0.54} & \textbf{\color{ForestGreen}0.64} & 0.35 & 0.47 \\
 & & 3 & 0.24 & \textbf{0.61} & \textbf{\color{ForestGreen}0.73} & 0.40 & 0.58 \\
\midrule

\rowcolor{tableShade}
\textsc{Soft Radial} & 500 & 0.5 & $-$0.01 & \textbf{\color{ForestGreen}0.23} & \textbf{0.22} & 0.20 & 0.19 \\
\rowcolor{tableShade}
 & & 1 & $-$0.01 & \textbf{0.37} & \textbf{\color{ForestGreen}0.41} & 0.33 & 0.36 \\
\rowcolor{tableShade}
 & & 2 & $-$0.01 & 0.51 & \textbf{\color{ForestGreen}0.60} & 0.47 & \textbf{0.52} \\
\rowcolor{tableShade}
 & & 3 & $-$0.01 & 0.58 & \textbf{\color{ForestGreen}0.68} & 0.54 & \textbf{0.61} \\
 & 1000 & 0.5 & $-$0.00 & \textbf{0.26} & \textbf{\color{ForestGreen}0.27} & 0.23 & 0.21 \\
 & & 1 & $-$0.00 & \textbf{0.40} & \textbf{\color{ForestGreen}0.45} & 0.36 & 0.38 \\
 & & 2 & $-$0.00 & \textbf{0.55} & \textbf{\color{ForestGreen}0.62} & 0.49 & \textbf{0.55} \\
 & & 3 & $-$0.00 & 0.62 & \textbf{\color{ForestGreen}0.70} & 0.56 & \textbf{0.63} \\
\rowcolor{tableShade}
 & 2500 & 0.5 & $-$0.00 & \textbf{0.28} & \textbf{\color{ForestGreen}0.30} & 0.25 & \textbf{0.28} \\
\rowcolor{tableShade}
 & & 1 & $-$0.00 & \textbf{0.43} & \textbf{\color{ForestGreen}0.47} & 0.38 & \textbf{0.43} \\
\rowcolor{tableShade}
 & & 2 & $-$0.00 & \textbf{0.59} & \textbf{\color{ForestGreen}0.64} & 0.51 & 0.55 \\
\rowcolor{tableShade}
 & & 3 & $-$0.00 & \textbf{0.66} & \textbf{\color{ForestGreen}0.72} & 0.58 & 0.65 \\
 & 5000 & 0.5 & $-$0.00 & \textbf{0.29} & \textbf{\color{ForestGreen}0.31} & 0.25 & \textbf{0.29} \\
 & & 1 & $-$0.00 & \textbf{0.45} & \textbf{\color{ForestGreen}0.48} & 0.39 & \textbf{0.45} \\
 & & 2 & $-$0.00 & \textbf{0.61} & \textbf{\color{ForestGreen}0.65} & 0.52 & 0.56 \\
 & & 3 & $-$0.00 & \textbf{0.69} & \textbf{\color{ForestGreen}0.73} & 0.59 & 0.62 \\
\end{longtable}
\end{small}

\vspace{-0.5em}
\noindent\textit{\small Notes: Each cell reports out-of-sample $R^2$. Models: OLS (Ordinary Least Squares), RF (Random Forest), MLP (Multi-Layer Perceptron), GBM (Gradient Boosting), Att Reg (Attention Regression, $M{=}5$ heads). \textbf{\color{ForestGreen}Green bold} = best; \textbf{black bold} = second-best.}

\section{Regression Block: Architecture}\label{app:regblk}

\paragraph{Skeleton.}
The Regression Block replaces the multi-head softmax attention sublayer of an FT-Transformer-style block \citep{gorishniy2021revisiting} with an explicit regression, and is used as a \emph{single} block. It keeps the FT-Transformer block's pattern, a mixing sublayer and a position-wise FFN, each wrapped in a residual connection with post-norm LayerNorm. Three skeleton details differ from the FT-Transformer baseline and are stated because they matter for reproduction. The tokenizer is a shared linear map $\mathbb{R}\to\mathbb{R}^{d}$ with a learned positional embedding per feature (the FT-Transformer uses a per-feature tokenizer). The output is a mean over tokens followed by a linear layer (the FT-Transformer uses a learned [CLS] token). The FFN activation is ReLU (GELU in the baseline).

\paragraph{Mixing sublayer.}
Inside the mixing sublayer, the token sequence $\boldsymbol{h}\in\mathbb{R}^{P\times d}$ is mapped to degree-2 polynomial cross-features per token, linear and squared terms plus all pairwise products across feature positions, $O(P^2)$ in width. These are compressed by a PCA with 200 components, fit once on the training tokens (at most 5{,}000 rows) and then frozen, and passed through a linear map $\boldsymbol{W}$ whose output is added back to the residual stream:
\[
\text{LN}\bigl(\boldsymbol{h}+\boldsymbol{W}\,\text{PCA}(\text{polycross}(\boldsymbol{h}))\bigr).
\]
There is no kernel and no learned similarity metric. Every coefficient is a regression coefficient. The PCA rotation is unsupervised, but it does not determine the predictor: PCA chooses the coordinate system and the compression, and supervised training chooses the target-relevant combination within that system.

\paragraph{Warm start.}
Before training, the first output channel of $\boldsymbol{W}$ and the head are warm-started from a ridge regression ($\lambda=1$) of the pooled PCA features on the training target, so training starts from ridge-based weights. This is the ridge embedding of Section~\ref{sec:reducdim} put to work.

\paragraph{Training.}
Adam at a constant learning rate of $10^{-3}$ with weight decay $10^{-4}$, batch $256$, dropout $0.1$, gradient clipping at norm $1$, and up to 200 epochs with early stopping (patience 10 on a 15\% validation split). The same configuration is used on every dataset. The FT-Transformer baseline shares the learning rate, weight decay, batch size, budget and early-stopping rule, and uses AdamW with a linear warm-up over the first 20 epochs and cosine decay.

\paragraph{Parameter count.}
With $d_{\text{model}}=64$ and $d_{\text{ff}}=128$, the model totals $30{,}401$ trainable parameters at $P=8$, against $101{,}697$ for the FT-Transformer baseline (three blocks, $d_{\text{model}}=64$, four heads, $d_{\text{ff}}=128$).

\section{Protocol and Uncertainty}\label{app:protocol}

Every method in Table~\ref{tab:realdata_main} runs under one convention: features and targets standardized on the training split, 5 random 80/20 splits, one fixed configuration per method with no per-dataset tuning, and a cap of $N=5000$ on the larger datasets (lifted in Appendix~\ref{app:scope}).

\paragraph{Paired comparisons.} Table~\ref{tab:se} reports the Regression Block minus the FT-Transformer, paired over the 5 splits of each dataset (two-sided paired $t$-test) and pooled over all 40 dataset$\times$split observations.

\begin{apptable}
\centering
\caption{Paired comparison of the Regression Block and the FT-Transformer.}
\label{tab:se}
\small
\begin{tabular}{@{}l ccc c c@{}}
\toprule
\textbf{Dataset} & \textbf{FT-T} & \textbf{Reg.~Block} & \textbf{Difference} & \textbf{SE} & \boldmath$p$ \\
\midrule
California & 0.764 & 0.766 & +0.0020 & 0.0062 & 0.76 \\
Yacht & 0.988 & 0.989 & +0.0013 & 0.0064 & 0.85 \\
Energy & 0.994 & 0.995 & +0.0002 & 0.0013 & 0.86 \\
Concrete & 0.898 & 0.907 & +0.0091 & 0.0071 & 0.27 \\
Airfoil & 0.924 & 0.927 & +0.0030 & 0.0104 & 0.78 \\
Abalone & 0.324 & 0.329 & +0.0054 & 0.0048 & 0.32 \\
Kin8nm & 0.914 & 0.920 & +0.0057 & 0.0017 & 0.03 \\
Protein & 0.426 & 0.455 & +0.0289 & 0.0088 & 0.03 \\
\midrule
Pooled (40 pairs) & & & +0.0070 & 0.0025 & 0.009 \\
\bottomrule
\end{tabular}
\vspace{0.1em}
\begin{minipage}{0.97\linewidth}
\footnotesize
\textit{Notes:} Regression Block minus FT-Transformer, out-of-sample $R^2$ under the protocol of Table~\ref{tab:realdata_main}, paired over the 5 splits of each dataset (SE and two-sided $p$ from a paired $t$-test) and pooled over all 40 dataset$\times$split observations. The Regression Block is ahead on 8/8 datasets (sign test $p=0.008$), and two of eight margins are individually significant at the 5\% level.
\end{minipage}
\end{apptable}

\section{Ablations}\label{app:ablations}

\paragraph{Design.}
The Regression Block changes several things at once relative to the FT-Transformer: the polynomial expansion, PCA compression, the removal of softmax attention, the parameter count, and, as Appendix~\ref{app:regblk} states, the depth, one block against three. The construction gives four testable predictions.

If attention is replaceable by an explicit nonlinear regression, the regression head and the polynomial expansion should both matter, since together they stand in for attention. PCA compression should be a convenience, a compression the low-rank view itself motivates, rather than a source of accuracy. The polynomial expansion should matter most where there is nonlinearity to capture, which the gap between Random Forest and least squares flags independently. And adding polynomial inputs to a standard FT-Transformer should be largely redundant, since attention already supplies the nonlinearity.

\paragraph{Configurations.}
Ten configurations, all on the datasets, splits, seeds, budget and protocol of Table~\ref{tab:realdata_main}.
\begin{itemize}\itemsep 0.1em
\item[A0] The FT-Transformer baseline (3 blocks, $d_{\text{model}}=64$).
\item[A1] A0 with degree-2 polynomial inputs added.
\item[A2] The Regression Block without the polynomial expansion.
\item[A3] The Regression Block without PCA.
\item[A4] Softmax attention put back into the Regression Block's own one-block skeleton, so the clean head comparison is A4 against RB.
\item[A5] The FT-Transformer shrunk to the Regression Block's budget, keeping its three blocks and narrowing $d_{\text{model}}$ (selected at 36 on every dataset).
\item[A6] The Regression Block stacked to three blocks.
\item[A7] The Regression Block with its single block widened to the FT-Transformer's budget.
\item[RB] The Regression Block.
\item[T] The trimmed configuration: two mixing sublayers, no feed-forward sublayer, 50 principal components.
\end{itemize}
Table~\ref{tab:abl} reports the grid, with means and parameter counts. All contrasts below are paired over the 40 dataset$\times$split observations.

\begin{apptable}
\centering
\caption{Ablation grid, out-of-sample $R^2$.}
\label{tab:abl}
\footnotesize
\setlength{\tabcolsep}{3pt}
\begin{tabular}{@{}l cccccccccc@{}}
\toprule
\textbf{Dataset} & \textbf{A0} & \textbf{A1} & \textbf{A2} & \textbf{A3} & \textbf{A4} & \textbf{A5} & \textbf{A6} & \textbf{A7} & \textbf{RB} & \textbf{T} \\
\midrule
California & 0.764 & 0.769 & 0.701 & 0.757 & 0.728 & 0.772 & 0.736 & 0.768 & 0.766 & 0.759 \\
Yacht & 0.988 & 0.981 & 0.572 & 0.990 & 0.538 & 0.982 & 0.968 & 0.989 & 0.989 & 0.989 \\
Energy & 0.994 & 0.994 & 0.941 & 0.977 & 0.922 & 0.980 & 0.991 & 0.995 & 0.995 & 0.988 \\
Concrete & 0.899 & 0.898 & 0.806 & 0.874 & 0.814 & 0.873 & 0.888 & 0.908 & 0.907 & 0.880 \\
Airfoil & 0.924 & 0.926 & 0.524 & 0.908 & 0.865 & 0.913 & 0.933 & 0.921 & 0.929 & 0.912 \\
Abalone & 0.324 & 0.314 & 0.286 & 0.324 & 0.328 & 0.324 & 0.310 & 0.330 & 0.329 & 0.327 \\
Kin8nm & 0.914 & 0.909 & 0.449 & 0.923 & 0.821 & 0.911 & 0.885 & 0.917 & 0.920 & 0.905 \\
Protein & 0.427 & 0.391 & 0.295 & 0.434 & 0.337 & 0.418 & 0.461 & 0.469 & 0.456 & 0.435 \\
\midrule
Mean & 0.779 & 0.773 & 0.572 & 0.773 & 0.669 & 0.772 & 0.772 & 0.787 & 0.786 & 0.774 \\
Params ($P=8$) & 101,697 & 106,305 & 21,697 & 173,249 & 34,177 & 33,013 & 89,921 & 101,249 & 30,401 & 7,489 \\
\bottomrule
\end{tabular}
\vspace{0.1em}
\begin{minipage}{0.97\linewidth}
\footnotesize
\textit{Notes:} Protocol of Table~\ref{tab:realdata_main}. The A0 and RB columns are independent reruns of the FT-T and Reg.~Block columns of Table~\ref{tab:realdata_main} and agree with them to within $0.002$ on every dataset. T is the trimmed configuration of the list above: two mixing sublayers, no feed-forward sublayer, 50 principal components. Parameter counts are at $P=8$.
\end{minipage}
\end{apptable}

\paragraph{Results.}
The results bear out the four predictions in their broad pattern, and each contrast below states its exceptions.

\textit{Polynomial expansion.} Removing it (A2) is the Regression Block's most costly change ($-0.215$, $p<0.001$), worst where Random Forest most outruns least squares: across the eight datasets the size of the loss tracks that gap with a rank correlation of $0.74$. Adding polynomial inputs to the FT-Transformer (A1) is largely redundant, and where they move most, they hurt (Protein $-0.036$).

\textit{Softmax head.} Softmax back in the same one-block skeleton (A4) loses $0.117$ ($p=0.003$) and is worse than the Regression Block on all eight datasets and worse than the FT-Transformer on seven. The head accounts for most of the performance difference within this controlled comparison.

\textit{PCA.} Removing it (A3) costs $0.013$ ($p<0.001$) at $5.7\times$ the parameters, so compression buys parameter economy, not accuracy. A3 is ahead of the Regression Block only on Yacht and marginally on Kin8nm.

\textit{Capacity.} The FT-Transformer shrunk to the Regression Block's budget (A5) loses $0.008$ ($p=0.006$) but does not break, softmax at three narrow blocks holding (mean $0.772$) where softmax at one block did not ($0.669$). The Regression Block widened to the FT-Transformer's budget (A7) changes nothing ($+0.001$, $p=0.57$). A5 and A7 together say that capacity does not explain the comparison in either direction.

\textit{Depth.} The Regression Block stacked to three blocks (A6), matching the FT-Transformer's depth at triple the parameters, gets worse ($-0.015$, $p<0.001$). Within these architectures softmax benefits substantially from depth, whereas the regression head performs best with one block. A4 and A6 say that depth is not a hidden advantage of the Regression Block. The results throughout use a single block, and I retain the one-block specification without claiming that the idempotence result of Appendix~\ref{app:stack} governs nonlinear deep Transformers.

\paragraph{How far the Regression Block can be cut.} Removing the feed-forward sublayer and truncating the principal-component basis from 200 to 50 leaves two mixing sublayers and a linear head, the second mixer preserving the Regression Block's two-sublayer depth once the feed-forward sublayer goes: $7{,}489$ parameters at $P=8$, against $30{,}401$ for the Regression Block and $101{,}697$ for the FT-Transformer, 7.4 percent of the latter. I report this as an ablation, not as a proposed architecture (Table~\ref{tab:abl}, column T). Paired over the 40 observations it is within $0.005$ of the FT-Transformer in mean $R^2$ ($0.774$ against $0.779$), ahead on three of eight, the difference small but signed against it ($p=0.047$). Against the Regression Block it gives up $0.011$ ($p<0.001$).

Separating the two cuts, 50 principal components already retain more than $99.99\%$ of the variance of the polynomial basis, so the basis width costs almost nothing (mean $R^2$ of $0.785$ at 200 components against $0.783$ at 50 in an independent rerun), while the feed-forward sublayer is worth about $0.009$. Appendix~\ref{app:scope} reports the trimmed configuration at higher dimension.

\section{Scope: Full Sample Size, Higher Dimensions, Classification}\label{app:scope}

\paragraph{Full sample size.}
The cap of $N=5000$ in Table~\ref{tab:realdata_main} binds on three of the eight datasets, and Panel A of Table~\ref{tab:scope_reg} lifts it. Across all eight datasets at full sample size the average rank among the four from-scratch methods is $1.62$ for the Regression Block, $2.12$ for the FT-Transformer, $2.25$ for Random Forest and $4.00$ for least squares. The two are close, the Regression Block ahead of the FT-Transformer on six of eight with a mean difference of $+0.003$. An explicit regression in the attention sublayer performs comparably at a fraction of the parameters.

\begin{apptable}
\centering
\caption{Scope of the regression comparison, out-of-sample $R^2$.}
\label{tab:scope_reg}
\small
\setlength{\tabcolsep}{4pt}
\begin{tabular}{@{}l r r cccc cc@{}}
\toprule
\textbf{Dataset} & \boldmath$N_{\text{train}}$ & \boldmath$P$ & \textbf{OLS} & \textbf{RF} & \textbf{FT-T} & \textbf{Reg.~Block} & \textbf{TabPFN} & \textbf{TabICL} \\
\midrule
\multicolumn{9}{@{}l}{\textit{Panel A: Full sample size}} \\
California & 16,512 & 8 & 0.601 & 0.817 & 0.807 & 0.805 & 0.872 & 0.878 \\
Kin8nm & 6,553 & 8 & 0.410 & 0.691 & 0.924 & 0.931 & 0.935 & 0.935 \\
Protein & 36,584 & 9 & 0.275 & 0.678 & 0.617 & 0.613 & 0.773 & 0.779 \\
\midrule
\multicolumn{9}{@{}l}{\textit{Panel B: Higher dimensions (capped at $N=5000$)}} \\
CPU\_act & 4,000 & 21 & 0.708 & 0.978 & 0.977 & 0.973 & 0.985 & 0.985 \\
Bank32nh & 4,000 & 32 & 0.526 & 0.503 & 0.535 & 0.509 & 0.564 & 0.562 \\
Ailerons & 4,000 & 40 & $<\!0$ & 0.827 & 0.844 & 0.834 & 0.867 & 0.865 \\
Pol & 4,000 & 48 & 0.488 & 0.977 & 0.993 & 0.991 & 0.994 & 0.989 \\
Superconductivity & 4,000 & 81 & 0.720 & 0.878 & 0.831 & 0.846 & 0.884 & 0.889 \\
\bottomrule
\end{tabular}
\vspace{0.1em}
\begin{minipage}{0.97\linewidth}
\footnotesize
\textit{Notes:} Protocol of Table~\ref{tab:realdata_main}. Panel A lifts the cap of $N=5000$ on the three datasets where it binds. Panel B applies the same method, unchanged and untuned, to datasets of increasing dimension, capped at $N=5000$. The Regression Block entry for Superconductivity averages four seeds (one run timed out). TabPFN and TabICL are pretrained tabular foundation models, included as reference points.
\end{minipage}
\end{apptable}

\paragraph{Higher dimensions.} Panel B of Table~\ref{tab:scope_reg} applies the same method, unchanged and untuned, to real regression datasets of increasing dimension (capped at $N=5000$, protocol of Table~\ref{tab:realdata_main}). The check concerns the from-scratch architectures: does the construction survive dimension without tuning? The Regression Block stays within $0.026$ of the FT-Transformer at every dimension tested, and is ahead of it on the highest-dimensional dataset, where Random Forest leads. The $O(P^2)$ cross-feature map is the practical constraint: at $P=81$ the Regression Block runs with small batches, and the trimmed configuration of Appendix~\ref{app:ablations}, with a fixed 50-component budget, loses $0.010$ to $0.036$ relative to the Regression Block, $0.012$ to $0.062$ relative to the FT-Transformer, at $P$ between 21 and 48 (CPU\_act $0.963$, Bank32nh $0.473$, Ailerons $0.817$, Pol $0.981$), whereas at $P\le 9$ it is within $0.005$ of the FT-Transformer.

\paragraph{Classification.} The Regression Block is adapted by a cross-entropy head and a multinomial-logistic warm start, otherwise unchanged, on six OpenML datasets (Table~\ref{tab:clf}). Among the methods trained from scratch it has the best average rank, $2.08$, against $2.17$ for Random Forest, $2.58$ for logistic regression and $3.17$ for the FT-Transformer (fractional ranks, ties split), and it is not the strongest scratch-trained method on every dataset. The architecture extends to classification by replacing the squared-error head with a cross-entropy head. A discrete outcome adds a link function, which the exact identity of Section~\ref{sec2} does not survive, and the regression reading then applies to the class scores.

\begin{apptable}
\centering
\caption{Classification accuracy on six OpenML datasets.}
\label{tab:clf}
\small
\begin{tabular}{@{}l cccc cc@{}}
\toprule
\textbf{Dataset} & \textbf{LogReg} & \textbf{RF} & \textbf{FT-T} & \textbf{Reg.~Block} & \textbf{TabPFN} & \textbf{TabICL} \\
\midrule
BreastCancer & 0.977 & 0.949 & 0.963 & 0.967 & 0.984 & 0.981 \\
Diabetes & 0.779 & 0.781 & 0.774 & 0.775 & 0.781 & 0.781 \\
Phoneme & 0.752 & 0.903 & 0.872 & 0.857 & 0.901 & 0.914 \\
Wine & 0.967 & 0.972 & 0.961 & 0.967 & 0.983 & 0.994 \\
Vehicle & 0.794 & 0.748 & 0.782 & 0.846 & 0.881 & 0.892 \\
Segment & 0.938 & 0.973 & 0.968 & 0.974 & 0.988 & 0.990 \\
\midrule
Mean & 0.868 & 0.888 & 0.887 & 0.898 & 0.920 & 0.925 \\
\bottomrule
\end{tabular}
\vspace{0.1em}
\begin{minipage}{0.97\linewidth}
\footnotesize
\textit{Notes:} Accuracy averaged over 5 splits. The Regression Block uses a cross-entropy head and a multinomial-logistic warm start, otherwise unchanged. TabPFN and TabICL are pretrained tabular foundation models, included as reference points.
\end{minipage}
\end{apptable}

\paragraph{Foundation models as reference points.} The two rightmost columns of Tables~\ref{tab:scope_reg} and~\ref{tab:clf} are the tabular foundation models TabPFN \citep[v6.4.1, full training sets, no subsampling]{hollmann2025tabpfn,grinsztajn2025tabpfn25} and TabICL \citep[regressor from the v2 release,][]{qu2025tabicl,qu2026tabiclv2}. They are reference points rather than competitors: production-grade systems pretrained on large synthetic corpora, whereas the test of the paper's hypothesis is the comparison with the FT-Transformer, the same architecture with and without the substitution. What the columns add is a sense of how far a Regression Block trained from scratch, with about 30K parameters and no tuning, sits behind such systems, and the distance is modest: on the eight benchmarks of Table~\ref{tab:realdata_main}, in runs under the same protocol reported in the repository, TabPFN averages $0.835$ against $0.786$ for the Regression Block, and the pattern in Tables~\ref{tab:scope_reg} and~\ref{tab:clf} is the same.

\section*{Impact Statement}

This paper presents work whose goal is to advance the field of Machine Learning by providing a theoretical bridge between classical ordinary least squares regression and modern attention mechanisms. The primary contribution is pedagogical and interpretive: making attention mechanisms more accessible to researchers and practitioners familiar with traditional statistics.

The potential positive societal impacts include improved model interpretability and pedagogical clarity for a broader audience. By revealing the structural alignment between least squares and attention, practitioners may develop better intuitions about how modern deep learning architectures process information.

I acknowledge the risk that oversimplifying or misapplying this analogy could lead to misconceptions about the capabilities or limitations of Transformer-based models. The equivalence shown is restricted and does not imply that large language models reduce to linear regression in practice. There are no direct ethical concerns or dual-use risks associated with this theoretical work.

\newpage
\section*{NeurIPS Paper Checklist}

\begin{enumerate}

\item {\bf Claims}
  \item[] Question: Do the main claims made in the abstract and introduction accurately reflect the paper's contributions and scope?
  \item[] Answer: \answerYes{}
    \item[] Justification: The abstract, the ``Double Life of Least Squares'' paragraph and the paragraph that follows it in the Introduction state the two contributions, the exact identity (Section~\ref{sec2}, Eq.~\eqref{ols:att}) and its two empirical uses (Section~\ref{sec:exp}), in the same terms as the body of the paper.

\item {\bf Limitations}
  \item[] Question: Does the paper discuss the limitations of the work performed by the authors?
  \item[] Answer: \answerYes{}
  \item[] Justification: The Limitations paragraph of Section~\ref{sec3} states the two limitations: the exact identity is a one-layer statement whose reading in a full Transformer is local to each module (Section~\ref{sec:depth}), and the Regression Block's cross-features grow quadratically with the number of inputs (Appendix~\ref{app:scope}). The introduction says that the paper does not claim that LLMs perform OLS.

\item {\bf Theory assumptions and proofs}
  \item[] Question: For each theoretical result, does the paper provide the full set of assumptions and a complete (and correct) proof?
  \item[] Answer: \answerYes{}
  \item[] Justification: All algebraic results (Eqs.~\eqref{ols_standard}--\eqref{ols:att} and the optimal-embedding solution Eq.~\eqref{ols_w}) are stated with their assumptions (full-rank $\boldsymbol{X}'\boldsymbol{X}$) and derived in the main text. The uniqueness of the optimal embedding is stated in Section~\ref{sec2} and proved in Appendix~\ref{app:unique} (for a single target the embedding is not identified, and the text says so); the principal-component embedding is derived from the PCR prediction rule in Section~\ref{sec:reducdim}; the composition result for linear stacks with residual connections is in Appendix~\ref{app:stack}.

\item {\bf Experimental result reproducibility}
  \item[] Question: Does the paper fully disclose all the information needed to reproduce the main experimental results of the paper to the extent that it affects the main claims and/or conclusions of the paper?
  \item[] Answer: \answerYes{}
  \item[] Justification: Section~\ref{sec:exp} and Appendices~\ref{sec:simulation}, \ref{app:regblk} and \ref{app:protocol} describe the Attention Regression and Regression Block architectures, hyperparameters, optimizer, splits (5 random 80/20 for real data; fixed test set of $J=1000$ for Monte Carlo with 10 replications), and baselines. Datasets are standard public UCI/OpenML benchmarks.

\item {\bf Open access to data and code}
  \item[] Question: Does the paper provide open access to the data and code, with sufficient instructions to faithfully reproduce the main experimental results, as described in supplemental material?
  \item[] Answer: \answerYes{}
  \item[] Justification: Code, the per-seed results behind every real-data table, and the table generators are at \url{https://github.com/philgoucou/ols-attention}. All datasets used (California, Yacht, Energy, Concrete, Airfoil, Abalone, Kin8nm, Protein) are publicly available from UCI/OpenML and the paper specifies their source.

\item {\bf Experimental setting/details}
  \item[] Question: Does the paper specify all the training and test details (e.g., data splits, hyperparameters, how they were chosen, type of optimizer, etc.) necessary to understand the results?
  \item[] Answer: \answerYes{}
  \item[] Justification: For Attention Regression, the loss, regularization ($\lambda=10^{-3}$), L-BFGS with early stopping, Cholesky parameterization, and number of heads ($M=5$) are stated in Appendix~\ref{sec:simulation}. For the Regression Block, the architecture (FT-Transformer skeleton with a degree-2 polynomial+PCA regression in place of softmax attention) and parameter count ($\sim$30K) are stated in Section~\ref{sec:exp}. The Regression Block's architecture, ridge warm start, optimizer, schedule and early stopping are stated in Appendix~\ref{app:regblk}.

\item {\bf Experiment statistical significance}
  \item[] Question: Does the paper report error bars suitably and correctly defined or other appropriate information about the statistical significance of the experiments?
  \item[] Answer: \answerYes{}
  \item[] Justification: Monte Carlo results are averaged over 10 replications with $\pm 1$ standard deviation error bars across the 16 conditions per DGP shown in Figure~\ref{fig:model_performance}. Real-data results are averaged over 5 random 80/20 splits; the Regression Block vs.\ FT-Transformer comparison reports per-dataset paired standard errors, a sign test and a pooled paired test over the 40 dataset$\times$split observations (Table~\ref{tab:se}), and every ablation contrast in Appendix~\ref{app:ablations} is a paired test over the same 40 observations.

\item {\bf Experiments compute resources}
  \item[] Question: For each experiment, does the paper provide sufficient information on the computer resources (type of compute workers, memory, time of execution) needed to reproduce the experiments?
  \item[] Answer: \answerYes{}
  \item[] Justification: The Monte Carlo runs on a single CPU machine, each replication in minutes. The real-data benchmarks, the ablation suite and the scope experiments of Appendices~\ref{app:protocol}--\ref{app:scope} (about 1,500 training runs in total) were run on cloud GPUs (NVIDIA T4/A10G, with A100-80GB for the $P\ge 40$ datasets, where the $O(P^2)$ cross-feature map requires small batches); a single run takes seconds to a few minutes and the whole suite a few GPU-hours.

\item {\bf Code of ethics}
  \item[] Question: Does the research conducted in the paper conform, in every respect, with the NeurIPS Code of Ethics?
  \item[] Answer: \answerYes{}
  \item[] Justification: The work is theoretical/methodological and uses only publicly available, non-personal tabular regression benchmarks. No human subjects, no PII, no high-risk deployment.

\item {\bf Broader impacts}
  \item[] Question: Does the paper discuss both potential positive societal impacts and negative societal impacts of the work performed?
  \item[] Answer: \answerYes{}
  \item[] Justification: The Impact Statement section discusses positive impacts (interpretability, pedagogical clarity) and the risk that the OLS--attention analogy may be oversimplified.

\item {\bf Safeguards}
  \item[] Question: Does the paper describe safeguards that have been put in place for responsible release of data or models that have a high risk for misuse (e.g., pretrained language models, image generators, or scraped datasets)?
  \item[] Answer: \answerNA{}
  \item[] Justification: The paper does not release pretrained language models, image generators, or scraped datasets. The Regression Block is a small ($\sim$30K-parameter) tabular regression model with no plausible misuse vector.

\item {\bf Licenses for existing assets}
  \item[] Question: Are the creators or original owners of assets (e.g., code, data, models) used in the paper, properly credited and are the license and terms of use explicitly mentioned and properly respected?
  \item[] Answer: \answerYes{}
  \item[] Justification: All datasets are cited (UCI ML Repository, OpenML) and used under their open-data terms. Baselines (Random Forest, Gradient Boosting, MLP, FT-Transformer) are referenced via their original publications.

\item {\bf New assets}
  \item[] Question: Are new assets introduced in the paper well documented and is the documentation provided alongside the assets?
  \item[] Answer: \answerYes{}
  \item[] Justification: The code is released under the MIT license, with documentation, at the repository above. The paper does not release new datasets.

\item {\bf Crowdsourcing and research with human subjects}
  \item[] Question: For crowdsourcing experiments and research with human subjects, does the paper include the full text of instructions given to participants and screenshots, if applicable, as well as details about compensation (if any)?
  \item[] Answer: \answerNA{}
  \item[] Justification: No crowdsourcing or human subjects.

\item {\bf Institutional review board (IRB) approvals or equivalent for research with human subjects}
  \item[] Question: Does the paper describe potential risks incurred by study participants, whether such risks were disclosed to the subjects, and whether Institutional Review Board (IRB) approvals (or an equivalent approval/review based on the requirements of your country or institution) were obtained?
  \item[] Answer: \answerNA{}
  \item[] Justification: No human-subjects research; IRB approval not applicable.

\item {\bf Declaration of LLM usage}
  \item[] Question: Does the paper describe the usage of LLMs if it is an important, original, or non-standard component of the core methods in this research?
  \item[] Answer: \answerNA{}
  \item[] Justification: LLMs are not a component of the method: the identity, the estimators and the Regression Block are the author's, and no LLM enters any computation reported in the paper. LLMs (Anthropic's Claude) were used as assistants for writing and editing the text, for writing and running the experiment code of Appendices~\ref{app:protocol}--\ref{app:scope} under the author's direction, and for checking the manuscript against the code; all results were verified by the author.

\end{enumerate}

\end{document}